\documentclass[times,twocolumn,final]{elsarticle}
\usepackage{flushend}
\usepackage{medima}
\usepackage{framed,multirow}
\usepackage[table,xcdraw]{xcolor}
\usepackage{multicol}
\usepackage{amssymb}
\usepackage{latexsym}
\usepackage{booktabs}
\usepackage{url}
\usepackage{xcolor}
\usepackage{graphicx}
\usepackage{hyperref}
\usepackage{amsmath}
\usepackage{colortbl,hhline}
\usepackage{array,eucal}
\usepackage{algorithm}
\usepackage{algpseudocode} 
\usepackage{bm}
\newcommand{\ie}{\textit{i}.\textit{e}.}
\newcommand{\eg}{\textit{e}.\textit{g}.}

\usepackage{pifont}
\newcommand{\myparagraph}[1]{\vspace{0.1em}\noindent\textbf{#1}}
\definecolor{Gray}{gray}{0.95}
\newcolumntype{g}{>{\columncolor{Gray}} p{0.96cm}}

\usepackage{arydshln}
\newcolumntype{I}{!{\vrule width 3pt}}           
\newlength\savewidth
\newcommand\shline{\noalign{\global\savewidth\arrayrulewidth
\global\arrayrulewidth 1.0pt}%
\hline
\noalign{\global\arrayrulewidth\savewidth}}
\usepackage{color}

\definecolor{mygreen}{HTML}{39b54a}
\newcommand{\reshl}[2]{
#1  \fontsize{9pt}{1em}\selectfont\color{mygreen}{$\uparrow$ \textbf{#2}}
}
\newcommand{\reshld}[2]{
#1 \fontsize{9pt}{1em}\selectfont\color{mygreen}{$\downarrow$ \textbf{#2}}
}
\definecolor{newcolor}{rgb}{.8,.349,.1}
\journal{Medical Image Analysis}
\begin{document}
\verso{Xixi Jiang \textit{et~al.}}
\begin{frontmatter}
\title{Labeled-to-Unlabeled Distribution Alignment for Partially-Supervised Multi-Organ Medical Image Segmentation}%
\author[1]{Xixi \snm{Jiang}}
\author[1]{Dong \snm{Zhang}}
\author[2]{Xiang \snm{Li}}
\author[2]{Kangyi \snm{Liu}}
\author[1]{Kwang-Ting \snm{Cheng}}
\author[2]{Xin \snm{Yang}\corref{cor1}
\cortext[cor1]{Corresponding author}}
\ead{xinyang2014@hust.edu.cn}
\address[1]{Department of Electronic and Computer Engineering, The Hong Kong University of Science and Technology, Hong Kong, China}
\address[2]{School of Electronic Information and Communications, Huazhong University of Science and Technology, Wuhan 430074, China.}

\received{1 May 2023}
\finalform{10 May 2023}
\accepted{13 May 2023}
\availableonline{15 May 2023}
\communicated{S. Sarkar}
\begin{abstract}
Partially-supervised multi-organ medical image segmentation aims to develop a unified semantic segmentation model by utilizing multiple partially-labeled datasets, with each dataset providing labels for a single class of organs. However, the limited availability of labeled foreground organs and the absence of supervision to distinguish unlabeled foreground organs from the background pose a significant challenge, which leads to a distribution mismatch between labeled and unlabeled pixels. Although existing pseudo-labeling methods can be employed to learn from both labeled and unlabeled pixels, they are prone to performance degradation in this task, as they rely on the assumption that labeled and unlabeled pixels have the same distribution. In this paper, to address the problem of distribution mismatch, we propose a labeled-to-unlabeled distribution alignment (LTUDA) framework that aligns feature distributions and enhances discriminative capability. Specifically, we introduce a cross-set data augmentation strategy, which performs region-level mixing between labeled and unlabeled organs to reduce distribution discrepancy and enrich the training set. Besides, we propose a prototype-based distribution alignment method that implicitly reduces intra-class variation and increases the separation between the unlabeled foreground and background. This can be achieved by encouraging consistency between the outputs of two prototype classifiers and a linear classifier. Extenvive experimental results on the AbdomenCT-1K dataset and a union of four benchmark datasets (including LiTS, MSD-Spleen, KiTS, and NIH82) demonstrate that our method outperforms the state-of-the-art partially-supervised methods by a considerable margin, and even surpasses the fully-supervised methods. The source code is publicly available at \href{https://github.com/xjiangmed/LTUDA}{LTUDA}.
\end{abstract}
\end{frontmatter}
\section{Introduction}
\label{intro}
\begin{figure*}[htb]
\centering
\setlength{\abovecaptionskip}{-0.25cm}
\includegraphics[width=1.0\textwidth]{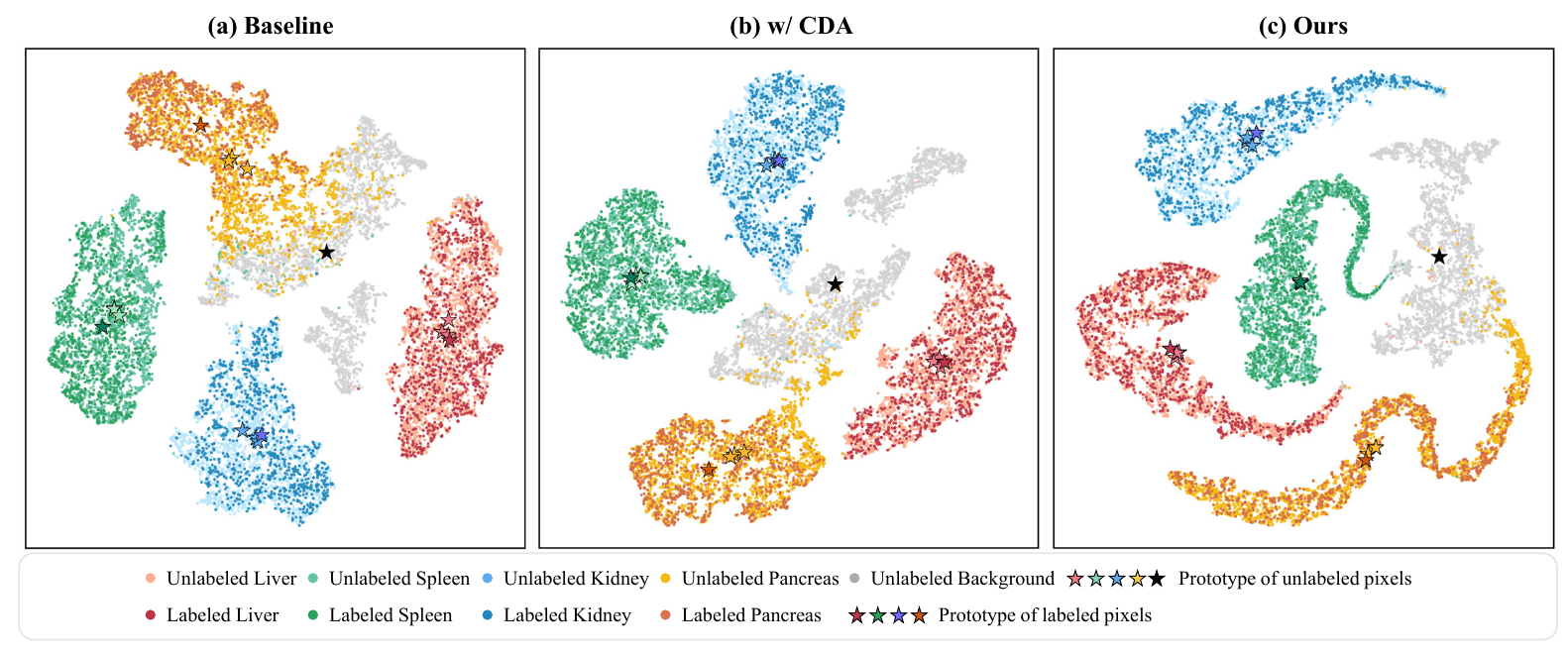}
\vspace{-2mm}
\caption{Comparisons of t-SNE feature visualization on the toy dataset consisting of four partially-labeled sub-datasets. The feature distribution of labeled and unlabeled pixels for different classes is visualized. For each foreground category, only one sub-dataset provides a labeled set, while the other three provide unlabeled sets. Since each sub-dataset does not provide the true label of the background, the background is completely unlabeled. We have superimposed the feature centers of the labeled set and unlabeled set, \ie, labeled prototypes and unlabeled prototypes, of each foreground category on the feature distribution. Additionally, we visualized the feature center of the background classes across all subsets. (a) Baseline model (trained on labeled pixels). The labeled prototype and unlabeled prototypes of the foreground classes are not aligned. (b) Baseline model with cross-set data augmentation (CDA). The CDA strategy effectively reduces the distributional discrepancy between labeled and unlabeled pixels for the foreground classes. (c) Our proposed method. The labeled prototype and unlabeled prototypes of each foreground class almost overlap.}
\vspace{-2mm}
\label{fig:schematic illustration}
\end{figure*}
Multi-organ medical image segmentation (Mo-MedISeg) is a fundamental yet challenging research task in the medical image analysis community. It involves assigning a semantic organ label, such as ``liver'', ``kidney'', ``spleen'', or ``pancreas'' to each pixel of a given image. Any undefined organ area is considered to belong to the ``background'' category~\citep{cerrolaza2019computational}. With the development of deep image processing technologies, such as convolutional neural networks (CNNs)~\citep{he2016deep} and vision transformers~\citep{dosovitskiy2020image,zhang2020feature,zhang2024cae}, Mo-MedISeg has gained significant research attention and has been widely applied in various practical applications, including diagnostic interventions~\citep{kumar2019multi} and treatment planning~\citep{chu2013multi}, as well as diverse scenarios, such as computed tomography (CT) images~\citep{gibson2018automatic} and X-ray images~\citep{gomez2020deep}.

However, training a fully-supervised deep semantic image segmentation model is challenging since it typically requires a large number of pixel-level labeled samples~\citep{wei2016stc,zhang2020causal}. This challenge is even more daunting for the Mo-MedISeg task, as obtaining accurate and dense multi-organ annotations is laborious, time-consuming, and requires scarce expert knowledge. Consequently, most benchmark public datasets, such as LiTS~\citep{bilic2023liver} and KiTS~\citep{heller2019kits19}, only provide annotations for one organ, with the remaining task-unrelated organs labeled as the background. Such partially annotated multi-organ datasets are more readily available compared to fully annotated multi-organ datasets. To alleviate the burden of collecting complete annotations, partially-supervised learning (PSL) has been used to learn Mo-MedISeg models from multiple partially annotated datasets~\citep{zhou2019prior,zhang2021dodnet,liu2022context}. Under this setting, each dataset provides labels for a single class of organs until all foreground categories of interest are covered. This strategy avoids the need for a densely labeled dataset and allows for the merging of datasets annotated with various types of organs from diverse institutions, especially when different hospitals focus on different organs.

A key challenge in using PSL for Mo-MedISeg is \emph{how to utilize limited labeled pixels and a large number of unlabeled pixels, without complicating the multi-organ segmentation model}. Each partially labeled dataset is annotated for a specific organ using a binary map that signifies whether a pixel belongs to the organ of interest. Labels for other foreground organs and background classes are not provided, resulting in a training set comprising both labeled and unlabeled pixels in each image. Prior methods typically learns solely from the labeled pixels~\citep{dmitriev2019learning,zhou2019prior,zhang2021dodnet}, or treats the missing labels as background~\citep{fang2020multi,shi2021marginal}. However, the lack of supervision for unlabeled structures can easily lead to an overfitting problem, while treating unlabeled anatomical structures as background can mislead and confuse the Mo-MedISeg model without prior information from fully annotated datasets. Although some advanced methods~\citep{huang2020multi,feng2021ms,liu2022context} have been proposed to generate pseudo-supervision for unlabeled pixels, these methods typically simply employ self-training to generate pseudo-labels. They use labeled pixels to train teacher models, \eg, multiple individual single-organ segmentation models for each dataset. The teacher model then generates pseudo-labels for unannotated organs in each partially-labeled dataset, resulting in a pseudo multi-organ dataset that is used to train the student multi-organ segmentation model. The performance of the student model depends on the quality of pseudo-labels. Unreliable pseudo-labels may lead to severe confirmation bias and performance degradation~\citep{chen2022debiased}.

This study aims to improve the quality of pseudo-labels for unlabeled pixels. An implicit assumption of pseudo-labeling methods is that labeled and unlabeled pixels follow the same distribution. In this paper, we contend that a \emph{distribution mismatch} occurs between the limited labeled pixels and the abundant unlabeled pixels, resulting in unreliable pseudo-labels. The origin of this distribution mismatch is twofold: 1) \emph{overfitting due to the limited labeled foreground pixels} and 2) \emph{spurious associations from a large number of unlabeled background pixels}. Fig.~\ref{fig:schematic illustration}(a) shows the feature visualization of the baseline model (More details of the baseline will be explained in section~\ref{Preliminary}), highlighting the aforementioned two issues. Firstly, the small sample size of labeled pixels causes the distribution of labeled pixels for foreground classes to deviate from the expected distribution of true samples, leading to the problem of overfitting. Secondly, the absence of supervision to distinguish between unlabeled foreground and background pixels results in the feature manifold of the background class being intertwined with that of the unlabeled foreground classes. The discriminative ability of the model is notably weakened, especially for organs (such as the pancreas) with low contrast to the background and indistinct boundaries. The distribution mismatch presents a significant challenge in accurately distinguishing the boundaries between organs and background. Applying the pseudo-labeling method directly to the PSL task cannot achieve satisfactory performance due to error propagation introduced by the biased pseudo-labels. 

To address the aforementioned issues, we propose a labeled-to-unlabeled distribution alignment (LTUDA) framework that reduces distribution bias and corrects pseudo-labels
from the distribution perspective. Our approach is based on two key insights. Firstly, unlabeled pixels can be used to guide data augmentation, which can broaden the distribution of training samples and alleviate overfitting. Secondly, compact representations can help to learn accurate decision boundaries and eliminate suspicious associations between unlabeled foreground and background categories. \emph{To address the first challenge}, we introduce a cross-set data augmentation strategy that generates augmented samples by interpolating between labeled and unlabeled pixels. The labels of the augmented samples are a combination of the ground truth of labeled pixels and pseudo-labels of unlabeled pixels. This cross-set mixing strategy leverages the advantages of both data augmentation and pseudo-labeling, enhancing the diversity of training samples. With the help of a large amount of unlabeled pixels, the distribution of labeled pixels can be approximated to that of unlabeled pixels, thereby reducing distribution bias. \emph{To address the second issue}, we propose a prototype-based distribution alignment (PDA) module that enhances the model's discriminative ability. The PDA module utilizes two non-parametric prototype classifiers to aid in the separation of ambiguous unlabeled pixels. Specifically, we enforce cross-set semantic consistency, where features from both the labeled and unlabeled pixels should be grouped together based on class similarity and separated from other classes. Well-clustered features can support the prototypes extracted from the labeled pixels to predict a good segmentation mask for the unlabeled pixels, and support the prototypes extracted from the unlabeled pixels to segment the labeled pixels accurately. To achieve ``labeled-to-unlabeled" and ``unlabeled-to-labeled" bidirectional segmentation, we enforce consistency between the outputs of the labeled prototype classifier and the unlabeled prototype classifier. This strategy bidirectionally aligns the distributions of labeled and unlabeled pixels, improving intra-class compactness and inter-class separability. 

Our main contributions are summarized as follows:
\begin{itemize}
  \item We highlight the distribution mismatch problem in partially-supervised medical image segmentation, and design a labeled-to-unlabeled distribution alignment framework to reduce distribution discrepancy, which in turn produces unbiased pseudo-labels.
  \item We present two effective strategies to reduce distribution bias. \emph{Firstly}, we propose cross-set data augmentation to bridge the distribution of labeled pixels and unlabeled pixels. \emph{Secondly}, we design a prototype-based distribution alignment method that implicitly facilitates learning aligned and compact feature representations.
  \item We demonstrate that our method can achieve superior performance to fully-supervised methods under the standard PSL setting and achieves state-of-the-art performance under different sizes of scarce labeled data. 
\end{itemize}
\section{Related work}
\subsection{Partially-supervised multi-organ segmentation}
 
An intuitive way of partially-supervised multi-organ segmentation is to train multiple independent segmentation models, one for each partially annotated dataset~\citep{isensee2021nnu,li2018h}. However, this approach has high computational costs and ignores the inter-organ relationship. To this end, condition-based methods~\citep{dmitriev2019learning,chen2019med3d,zhang2021dodnet} are proposed to train a single segmentation network, which encodes each segmentation task as a task-aware prior and adjusts network parameters (intermediate activation signals~\citep{dmitriev2019learning}, task-specific decoders~\citep{chen2019med3d} or dynamic segmentation heads~\citep{zhang2021dodnet}) to segment task-relevant organs. However, condition-based methods require multiple rounds of inference to obtain multi-organ segmentation results, which is time-consuming. 
Recently, several studies~\citep{zhou2019prior,fang2020multi,shi2021marginal,liu2022context} have attempted to develop a multi-organ segmentation model that can segment multiple organs with a single round of inference. Zhou et al.~\citep{zhou2019prior} 
incorporated anatomical priors on organ sizes from fully labeled datasets to regularize organ size distributions on partially labeled datasets. Fang et al.~\citep{fang2020multi} proposed a target adaptive loss that considers the unlabeled organs as background. Shi et al.~\citep{shi2021marginal} proposed a marginal loss to merge unlabeled organs with the background and an exclusion loss to impose mutual exclusiveness between different organs. Liu et al.~\citep{liu2022context} introduced a context-aware voxel-wise contrastive learning method to exploit unlabeled pixels. It is worth noting that these methods require a fully annotated dataset for training, which may not be feasible in many medical scenarios due to the challenges of obtaining fully annotated data. Other researchers~\citep{huang2020multi,feng2021ms,dong2022towards,zhang2022deep} have focused on learning solely from partially labeled datasets. Huang et al.~\citep{huang2020multi} and Feng et al.~\citep{feng2021ms} pre-trained multiple single-organ models to generate pseudo-labels for unlabeled organs, and then used the pseudo-full labels as supervision for the multi-organ segmentation model.  Dong et al.~\citep{dong2022towards} leveraged the similarity in human anatomy to generate vicinal labels by linearly combining randomly sampled partial labels. Zhang et al.~\citep{zhang2022deep} formulated partially-supervised learning as an optimization problem, and proposed conditional compatibility and dual compatibility to provide supervision for missing labels.

In this work, we aim to develop a multi-organ segmentation network from a union of partially annotated datasets. Our method overcomes limitations of existing methods in the following three aspects: (1) A general and efficient multi-organ segmentation model is proposed to overcome the limitations of existing methods, including the single function of the model, high training cost, and low testing efficiency; (2) The proposed method does not require fully labeled data; (3) Existing pseudo-labeling methods~\citep{huang2020multi,feng2021ms} for partially-supervised multi-organ segmentation do not take into account the distribution mismatch between labeled and unlabeled pixels, resulting in biased pseudo-labels. We focus on addressing the distribution mismatch problem, and propose a novel method to align their feature distributions, thereby correcting pseudo-labels from a distributional perspective.

\subsection{Partially-supervised learning}
In the partially-supervised multi-label classification task, the annotations are given in the form of partial labels. We collectively refer to this task and the partially-supervised multi-class segmentation task as partially-supervised learning. Only a subset of the categories are annotated in each image, so the treatment of unlabeled classes significantly affects the model learning. Existing methods in this field take one of three approaches to handle unlabeled classes:
\textbf{(1)~Ignore unlabeled categories.} The basic way to handle unlabeled categories is to ignore them~\citep{dmitriev2019learning,chen2019med3d,zhang2021dodnet,zhou2019prior}. With this training strategy, the model is trained solely on clean labels and will not be disturbed by incorrect labels. However, this approach has the drawback of using only a fraction of the available training labels, which may lead to overfitting and the learning of a weak representation.
\textbf{(2)~Treat unlabeled categories as negative.} Another way is to treat unknown labels as negative/ background~\citep{sun2010multi,kim2022large,fang2020multi,shi2021marginal}. In this approach, the unlabeled categories gain supervision. However, using such noisy labels can inevitably introduce false negative errors, which can negatively impact model performance. To address this issue, a recent study~\citep{kim2022large} has incorporated sample selection and label correction techniques to deal with label noise.
\textbf{(3)~Pseudo-supervision.} The third way is to generate pseudo-supervision for the unknown labels. Ignoring unlabeled classes can result in insufficient supervision signals, and recent works~\citep{durand2019learning,chen2022structured,abdelfattah2022plmcl,zhang2022effective,luo2020deep,kang2021label,verelst2023spatial,huang2020multi,feng2021ms} have used deep learning models to predict pseudo-labels for the unknown categories. In this way, partial labels of labeled data and pseudo-labels of unknown data can be used together to train the model, thus transforming PSL into fully-supervised learning. Following this line, we use the model's prediction to supplement the missing labels. 

\subsection{Learning from labeled and unlabeled data}
\emph{Self-training} and \emph{Consistency regularization} are two commonly used methods for training segmentation models with labeled and unlabeled data. 
\emph{Self-training} methods first train the model on labeled data, generate pseudo labels for unlabeled data, and then retrain the model on the combined labeled and pseudo-labeled data. The teacher and student models can be trained iteratively. Various schemes are introduced on how to generate reliable pseudo-segmentation maps~\citep{nair2020exploring,cao2020uncertainty,wang2021uncertainty,yang2022st++,zhang2020causal}. For example, selecting reliable pseudo-labels based on confidence or uncertainty estimation. Images or pixels with lower confidence will be re-weighted or removed from training.
\emph{Consistency regularization} encourages the model to have similar outputs for the same images under different perturbations, where perturbations are introduced in the input space~\citep{french2019semi,kim2020structured} or feature space~\citep{ouali2020semi,ke2020guided}. To enforce the decision boundary to lie in the low-density region, input perturbation methods impose a consistency constraint between the predictions of different randomly augmented images. On the other hand, feature perturbation employs a feature perturbation scheme by utilizing multiple decoders and enforcing consistency between their outputs.
Recently, researchers~\citep{sohn2020fixmatch,zou2020pseudoseg,chen2021semi,luo2022scribble,wu2023compete,chen2023confidence} are turning their attention to incorporate consistency regularization with self-training. Particularly, Fixmatch~\citep{sohn2020fixmatch} has been widely utilized for semi-supervised segmentation tasks~\citep{zou2020pseudoseg,seibold2022reference,wei2023segmatch}. The Fixmatch mechanism supervises a strongly augmented image $X_s$ with the prediction yielded from the corresponding weakly augmented image $X_w$. The idea behind this is that the model is more likely to produce a higher-quality prediction on $X_w$ than $X_s$. The strongly perturbed image $X_s$ can alleviate the confirmation bias caused by the inaccurate pseudo labels. In this study, we found that designing appropriate weak-to-strong consistency regularization can boost the performance for partially-supervised segmentation.

\begin{figure*}[t]
\centering
\setlength{\abovecaptionskip}{-0.25cm}
\includegraphics[width=\textwidth]{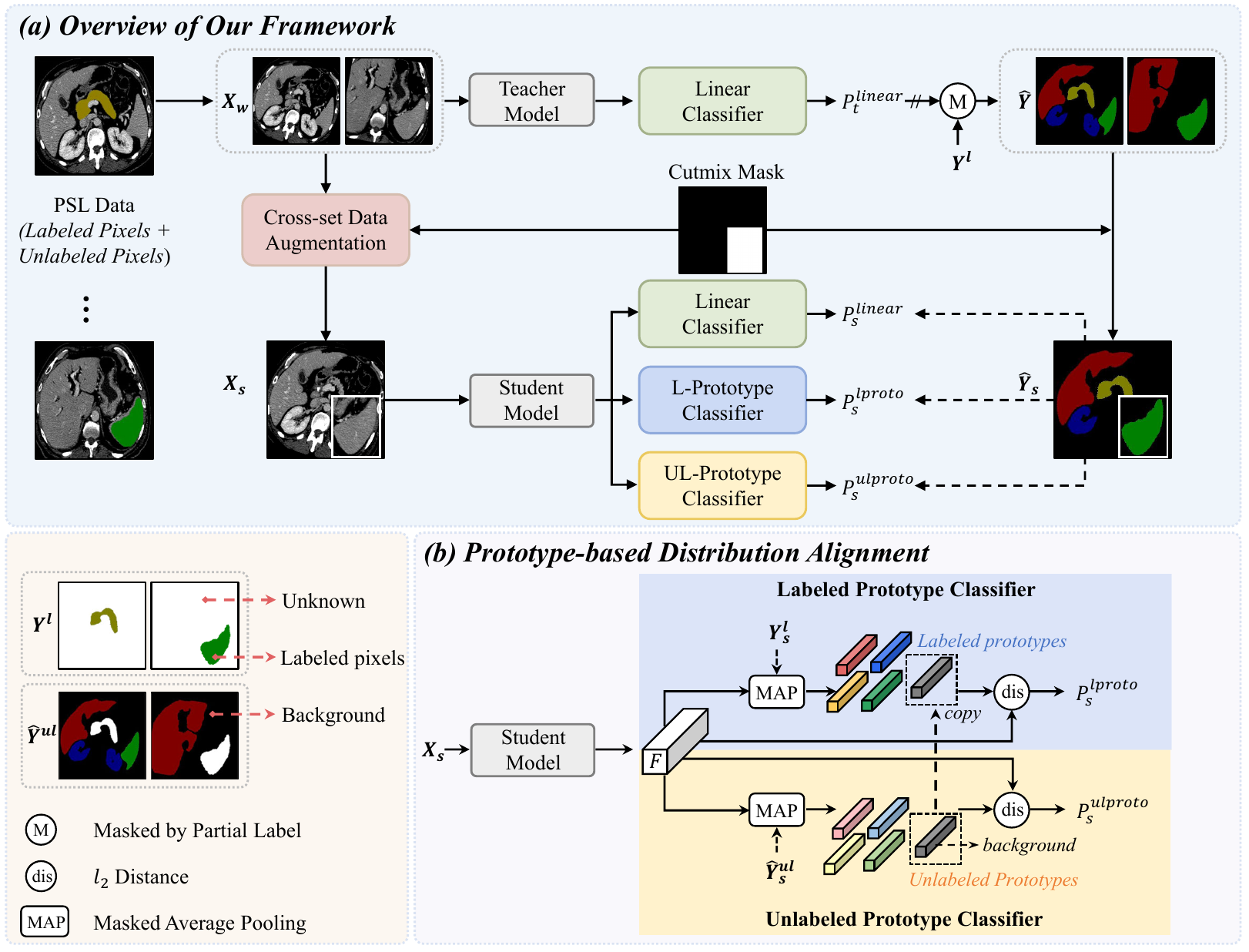}
\vspace{-2mm}
\caption{(a) The overall framework of the proposed LTUDA method, which consists of cross-set data augmentation and prototype-based distribution alignment. (b) Details of the prototype-based distribution alignment module. Our method is built on the popular teacher-student framework and applies weak (rotation and scaling) and strong augmentation (cross-set region-level mixing) to the input images of the teacher and student models, respectively. The linear classifier refers to the linear threshold-based classifier described in Equation~\ref{eq:threshold-based classifier} of Section~\ref{Preliminary}. Two prototype classifiers are introduced in the student model, and the predictions of the teacher model and partial labels are combined as pseudo-labels to supervise the outputs of the three classifiers in the student model. The term ``copy" denotes that the labeled prototype of the background class is set to be equal to the unlabeled prototype.}
	\label{framework}
\end{figure*}
\section{Methodology}
\subsection{Preliminaries}
\label{Preliminary}
Given a union of partially annotated datasets \{$D_1,D_2,...,D_C$\} ($D_c$=\{($X$,$Y^l$) $\mid$ $c$=1,2,...,$C$\}), our goal is to train an end-to-end segmentation network that can simultaneously segment $C$ organs. $X$ denotes the images in the $c$-th dataset. The corresponding partial label $Y^l$ comprises labeled foreground organ pixels, unlabeled foreground organ pixels, and unlabeled background pixels. In $Y^l$, only organ $c$ has true class labels, and other organs are unknown. The multi-organ segmentation network needs to assign each pixel of an input image a unique semantic label of $c \in \{0, 1, ..., C \}$, where $C$ denotes the class size. 

CNN-based Mo-MedISeg networks are commonly trained by multi-class cross-entropy loss $\mathcal{L}_{CE}(x,y) = -\sum_{c=0}^C y_c\log p_c$, where $y_c$ is the one-hot ground truth corresponding to class $c$. However, this is not suitable for partially-supervised segmentation. Given the absence of an annotated background class in this scenario, we learn class-specific foreground maps, and transform the multi-classification task into multiple binary classification tasks (one vs. the rest). Given an image from any partially labeled dataset $D_i$ as input, the segmentation network generates class-wise foreground maps $p=\{p_c\}_{c=1}^{C}$, which are normalized using a sigmoid. Partial binary cross-entropy loss $\mathcal{L}_{pBCE}$ is adopted to train the segmentation network, which is defined as follows:
\begin{equation}
    \mathcal{L}_{pBCE}(x,y)= \sum_{c=1}^C \mathbb{I}_{[y_c\neq -1]} (-y_c\log p_c-(1-y_c)\log(1-p_c)),
    \label{eq:pCE}
\end{equation}
where $y_c$ and $p_c$ represent the ground truth and predicted probability map belonging to class c, respectively. For the labeled category $i$, the value of $y_c$ is 0 or 1, indicating whether or not the pixel belongs to category $i$; for the unlabeled categories, the value of $y_c$ is -1.  $\mathcal{L}_{pBCE}$ only supervises the segmentation map corresponding to the labeled category. For example, the LiTs dataset provides partial labels for the liver category. Therefore, for some other categories without labels, this loss function will not back-propagate the gradients.

The segmentation network only outputs foreground maps, so we need to assign each pixel its unique class $\hat{y}$ from the label space $\{0,1,..., C \}$ during inference. Inspired by methods for detecting anomalous samples ~\citep{hendrycks2016baseline}, we adopt a linear threshold-based classifier to obtain multi-class segmentation maps. As shown in Equation~\ref{eq:threshold-based classifier}, if the probabilities of all foreground classes of a pixel are below a certain threshold $\tau$, we consider the pixel to belong to the background class. Otherwise, the pixel is assigned the class with the highest foreground probability.
\begin{equation}
\hat{y} =
\left\{
   \begin{array}{lr}
   \mathop{\arg\max}_{c \in \{1,\cdots, C\}}p_c, & \text{if } \max_{c \in \{1,\cdots,C\}}  p_c \ge \tau, \\
   0 \textrm{ (background class)}, & \text{otherwise}.
   \end{array}
\right.
\label{eq:threshold-based classifier}
\end{equation}
Although ignoring unannotated categories is an intuitive approach, it may lead to sub-optimal performance as only a fraction of the data are utilized. Fig.~\ref{fig:schematic illustration}(a) shows the feature visualization of the baseline model which only learns from the labeled pixels while ignoring the unlabeled organs. We can observe that the unlabeled prototypes deviate from the labeled prototype for the foreground categories, especially for the pancreas and spleen. Furthermore, it is difficult to separate the unlabeled foreground from the background only by the threshold $\tau$, due to the absence of any background supervision information. Therefore, directly minimizing $\mathcal{L}_{pBCE}$ on labeled pixels may result in the problems of overfitting and confusion between the foreground organs and the background. 

\subsection{Overall framework}
Fig.~\ref{framework} provides an overview of our proposed framework, which builds on the popular teacher-student framework Fixmatch~\citep{sohn2020fixmatch}, where the teacher model is obtained by applying the exponential moving average to the student model at each training iteration. The training batch consists of randomly sampled images from different partially labeled datasets. Random rotation and scaling are applied to the input images to obtain weakly augmented images $X_w$, and then cross-set data augmentation (CutMix~\citep{yun2019cutmix}) is applied to $X_w$ to obtain strongly augmented images $X_s$. To learn from partial labels, the student model has two objectives during training: 1) For the labeled pixels, the predictions of different augmented images $X_w$ and $X_s$ are expected to be close to the ground truth $Y^l$. (Note that we omit the part about using partial labels $Y^l$ to supervise the labeled pixels of the weak view $X_w$ in Fig.~\ref{framework}.)  2) For the unlabeled pixels, the predictions are expected to be close to the pseudo-labels $\hat{Y}^{ul}$. Online pseudo-labels are generated based on the teacher model's predictions for the weakly augmented images $X_w$, providing substantial supervision for the unlabeled pixels of the strongly augmented images $X_s$. Specifically, the weak images $X_w$ are fed into the teacher model backbone and linear classifier to get segmentation predictions $P_t^{linear}$, which are then transformed into hard pseudo-labels $\hat{Y}_t^{linear}$ using Equation~\ref{eq:threshold-based classifier}. Since the partial labels $Y^l$ provide the annotation information of certain organs, we use $Y^l$ to replace the predictions of the corresponding organs in $\hat{Y}_t^{linear}$ to obtain the masked pseudo-labels $\hat{Y}$. The strong images $X_s$ are fed into the student model, and the prediction results $P_s^{linear}$, $P_s^{lproto}$, $P_s^{ulproto}$ of the three branches are obtained through the linear classifier, labeled prototype classifier and unlabeled prototype classifier, respectively. Corresponding to the strong augmentation of the input image, we use CutMix to augment the pseudo-labels. Then the prediction results of the three branches are supervised by these augmented pseudo-labels $\hat{Y}_s$. The overall training loss for our framework can be formulated as follows:
\begin{eqnarray}
    \mathcal{L} &=& \mathcal{L}_{labeled}+ \mathcal{L}_{unlabeled} \nonumber \\
    &=& \mathcal{L}(X_w,Y^l)+\mathcal{L}(X_s,Y_s^l)+\mathcal{L}(X_s,\hat{Y}_s^{ul}) \nonumber \\  
    &=& \mathcal{L}_{pBCE}(X_w,Y^l)+ \mathcal{L}(X_s,\hat{Y}_s),
    \label{eq:total loss}
\end{eqnarray}
where $Y_s^l$ and $\hat{Y}_s^{ul}$ denote the strongly augmented versions of $Y^l$ and $\hat{Y}^{ul}$, respectively. $\hat{Y}_s$ is a combination of $Y_s^l$ and $\hat{Y}_s^{ul}$, so we combine $\mathcal{L}(X_s,Y_s^l)$ and $\mathcal{L}(X_s,\hat{Y}_s^{ul})$ into $\mathcal{L}(X_s,\hat{Y}_s)$. We will introduce the details of $\mathcal{L}(X_s,\hat{Y}_s)$ in Section~\ref{PDA}.

The main contribution of our method is that 1) we use cross-set strong data augmentation, performing mixing between labeled and unlabeled pixels; 2) We introduce two additional prototype classifiers in the student model, in addition to employing linear classifiers in both the teacher and student models. We utilize the pseudo-labels from the weak images to simultaneously constrain the segmentation results of the two prototype classifiers to achieve distribution alignment. Details will be explained in the subsequent sections.
\begin{figure}[tbp]
\centering
\setlength{\abovecaptionskip}{-0.25cm}
\includegraphics[width=0.5\textwidth]{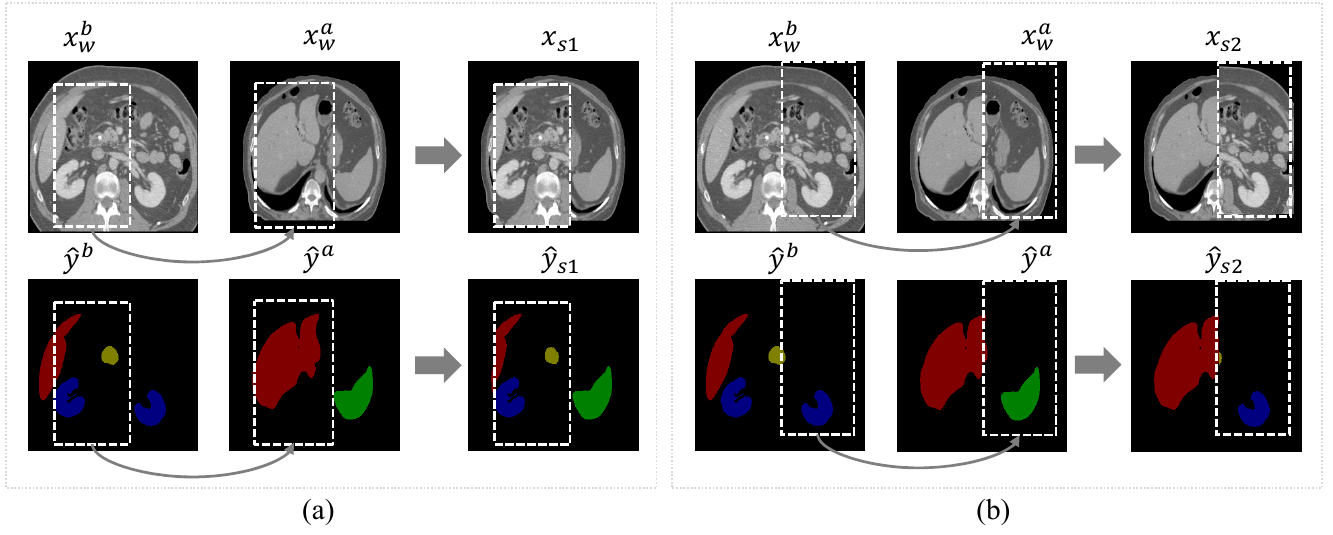}
\vspace{-2mm}
\caption{Visualizations of differently strongly augmented images generated by CutMix. (a) and (b) paste the cropped patch from $x_w^b$ to the same position in $x_w^a$. The box coordinates and sizes of (a) and (b) are different.}
\label{fig:cutmix_strong}
\end{figure}

\subsection{Cross-set data augmentation}
To address the overfitting problem arising from limited labeled pixels, we propose a cross-set mixing method to mitigate the impact of deviation in the distribution of labeled pixels by utilizing unlabeled pixels to guide the data augmentation process. Inspired by existing cross-domain mixing methods~\citep{xu2020adversarial,tranheden2021dacs,zhu2023progressive}, we use a region-level mixing method to combine labeled and unlabeled pixels via convex interpolation, generating new and strongly perturbed samples between them. Specifically, the PSL training set is augmented using CutMix~\citep{yun2019cutmix}, where a region of an image is replaced with the corresponding patch of another image from the training set.  Taking two images,  $x^a$, $x^b$, randomly sampled from the training batch as an example, the weakly augmented images are defined as $x^a_w = \mathcal{T}(x^a), {x^b_w} = \mathcal{T}(x^b)$, where $\mathcal{T}$ denotes the weak augmentation function (rotation and scaling). The masked pseudo-labels corresponding to $x^a_w$ and $x^b_w$ are denoted as $\hat{y}^a$ and $\hat{y}^b$, respectively. The strongly augmented image-label pair is denoted as:
\begin{equation}
    {x_s} = (1-M)\odot {x^a_w} + M\odot {x^b_w},
    \label{eq:strong_img}
\end{equation}
\begin{equation}
    {\hat{y}_s} = (1-M)\odot {\hat{y}^a} + M\odot {\hat{y}^b},
    \label{eq:strong_label}
\end{equation}
where $M$ is a binary mask representing the cut region of the image ${x^b_w}$. To obtain the binary mask $M$, we first sample a rectangular box $B=(r_x,r_y,r_w,r_h)$, where $r_x, r_y$, and $r_w,r_h$ represent the coordinates of the starting point, as well as the box sizes respectively. As shown in Equation~\ref{eq:sampling}, the box coordinates $r_x, r_y$ are uniformly sampled, and the box size $r_w,r_h$ is proportional to the size of the original image $W, H$. The ratio of the cropped area is $\frac{r_w r_h}{WH} = 1-\lambda$.
\begin{equation}
\begin{split}
    r_{x} \sim \text{Unif}~ \left(0, W\right), &~~~r_{w} = W \sqrt{1-\lambda},  \\
    r_{y} \sim \text{Unif}~ \left(0, H\right), &~~~r_{h} = H \sqrt{1-\lambda}.  \\
\end{split}
\label{eq:sampling}
\end{equation}
Then, region $B$ in $x_w^a$ is removed and filled with patches cropped from $B$ at the same location of $x_w^b$. 
Note that we do not restrict the coordinates of the rectangular box to fall within the partial label. A patch cropped from $x_w^a$ may contain both labeled pixels and unlabeled pixels.

The annotation of the generated samples combines the ground truth of the labeled pixels and the pseudo-labels of the unlabeled pixels. Using unlabeled pixels for data augmentation enriches the diversity of the augmented samples. At the same time, the noise of the pseudo-labels produced by biased predictions can be mitigated by mixing ground-truth labels. Therefore, the advantages of pseudo-labeling and data augmentation methods are drawn together in our method. Other region-based data mixing methods such as 
CowMix~\citep{french2020milking} and CarveMix~\citep{zhang2021carvemix} can also be used as cross-set data augmentation. 

Cross-set mixing serves as a bridge between labeled and unlabeled pixels, filling the gap between them by generating intermediate samples. Therefore, the distribution of the augmented data is closer to the true distribution than that of the original labeled pixels. Fortunately, region-level mixing can destroy the inherent structure of the original image, acting as a strong input perturbation. Using ${\hat{y}_s}$ to supervise $x_s$ can be seen as a combination of supervised and unsupervised consistency regularization, which helps to push the decision boundary to low-density regions.

Maximizing the consistency between different strongly perturbed views of unlabeled pixels allows for better utilization of the perturbation space ~\citep{berthelot2019remixmatch,caron2020unsupervised}. Inspired by this idea, we construct $M$ strong views (obtained by randomly applying strong augmentation $M$ times) to fully explore strong perturbations. A strongly augmented image is generated through a single-cut operation. The coordinates and box size of the cutting patch determine the disparity between different strongly augmented images. We show different strongly augmented images generated by CutMix in Fig.~\ref{fig:cutmix_strong}. We supervise the $M$ strongly augmented images ${ \{X_s^m\}}_{m=1}^M$ using masked pseudo-labels. The pseudo-supervision loss of $\mathcal{L}(X_s,\hat{Y}_s)$ in Equation~\ref{eq:total loss} can be calculated between each pair of one weak and one strong view. 

\begin{algorithm}[t]
    \small
    \caption{Training procedure}\label{alg:ours}
    \begin{algorithmic}
    \vspace{0.5mm}
    \hspace{-8mm}\colorbox[gray]{0.96}{
    \begin{minipage}{0.94\columnwidth}
    \State{\textbf{\textsc{Stage1: cross-set data augmentation}}}
    \State{\textbf{Input:} Training data $\{D_1,D_2,...,D_C\}$}
    \State{\textbf{Output:} Student model $\Theta$, teacher model $\widetilde{\Theta}$}
    \vspace{1mm}
    \State{Initialize $\Theta$ $\widetilde{\Theta}$ parameters randomly}
    \For{$t = 0,1,\dots, T_1-1$}
        \vspace{1mm}
        \State{Apply weak and strong data augmentation to obtain $X_w, X_s$}
        \State{Predict the missing labels for $X_w$ via $\widetilde{\Theta}$, and obtain $\hat{Y}_s$}
        \State{Estimate predictions for $X_w$, $X_s$ by $\Theta$ with linear classifier} 
        \State{Update $\Theta$ using $\mathcal{L}_{pBCE}(X_w,Y^l)+\mathcal{L}_{pBCE}(X_s,\hat{Y}_s;P_s^{linear})$} 
        \State{Update $\widetilde{\Theta}$ by $\Theta$ with moving-average} 
        \vspace{1mm}
    \EndFor 
\end{minipage}
}
        
    \vspace{1mm} 
    \hspace{-8mm}\colorbox[gray]{0.96}{
    \begin{minipage}{0.94\columnwidth}
        \State \textbf{\textsc{Stage2: prototype-based distribution alignment}}
        \State \textbf{Input:} Training data $\{D_1,D_2,...,D_C\}$, student model $\Theta$
        \State \textbf{Output:} Student model $\Theta$, teacher model $\widetilde{\Theta}$
        \vspace{1mm}
        \State Initialize $\Theta$, $\widetilde{\Theta}$ with the student parameters trained in \textbf{\textsc{Stage1}}
        \State $\mathcal{P}_{lproto} \gets 0$, $\mathcal{P}_{ulproto} \gets 0 $ 
        \For{$t = 0, 1 \dots, T_2-1$}
            \State Apply weak and strong data augmentation to obtain $X_w, X_s$
            \State Predict the missing labels for $X_w$ via $\widetilde{\Theta}$, and obtain $\hat{Y}_s$ 
            \State Estimate predictions for $X_w$, $X_s$ by $\Theta$ with linear classifier
            \State{Estimate predictions for $X_s$ by $\Theta$ with prototype classifiers}
            \State Update $\Theta$ using equation~\ref{eq:total loss} 
            \State Update prototypes $\mathcal{P}_{lproto}$, $\mathcal{P}_{ulproto}$ 
            \State Update $\widetilde{\Theta}$ by $\Theta$ with moving-average 
        \EndFor
    \end{minipage}
    }
 \end{algorithmic}
\end{algorithm}
\subsection{Prototype-based distribution alignment}
\label{PDA}
 It is difficult for a linear classifier to separate different foregrounds from the background through a fixed threshold $\tau$~\citep{zhou2021learning}. Considering that the features of the same class have semantic similarity, a prototype-based classifier can estimate the semantic segmentation results by measuring the distance between each pixel and the average feature (\ie, prototype) of different categories. We thus propose prototype-based distribution alignment to calibrate confusing features with the help of two prototype classifiers.
 
The ideal feature distribution ensures the features of labeled pixels and unlabeled pixels in the same category are closely grouped, while a substantial separation is maintained from those of other categories~\citep{guo2021semantic,wu2022exploring}. Moreover, a high-quality feature should empower any classifiers to make accurate predictions~\citep{oh2022daso}. When the distributions of labeled and unlabeled pixels align perfectly, the multi-class segmentation results based on the prototypes of labeled pixels should be comparable to the segmentation results using the prototypes of unlabeled pixels~\citep{wang2019panet,xu2022all}. Therefore, we apply consistency regularization to the predictions of different classifiers to implicitly align feature representations. Specifically, as shown in Fig.~\ref{framework}, we use the student model backbone to extract the features of strongly-augmented images $X_s$, and then calculate the prototypes for each category of the labeled and unlabeled pixels. For the labeled prototype classifier, the features of each category are aggregated according to the partial ground truth $Y^l_s$ to obtain the prototypes $\mathcal{P}_{lproto}$. Since there is no ground truth for the background class, we utilize pseudo-labels $\hat{Y}^{ul}_s$ to calculate its prototype. The segmentation result $P_s^{lproto}$ is obtained by calculating the distance from each pixel to the labeled prototypes. Conversely, we calculate the prototypes $\mathcal{P}_{ulproto}$ of the unlabeled pixels based on the pseudo-labels $\hat{Y}^{ul}_s$, and then perform segmentation with the unlabeled prototypes to get the prediction $P_s^{ulproto}$. Masked pseudo-labels are employed to simultaneously constrain the segmentation predictions of the linear classifier, labeled prototype classifier, and unlabeled prototype classifier. Thus, $\mathcal{L}(X_s,\hat{Y}_s)$ in equation~\ref{eq:total loss} is defined as
\begin{align}
    & \mathcal{L}(X_s,\hat{Y}_s) = \mathcal{L}_{linear}+ \mathcal{L}_{lproto} + \mathcal{L}_{ulproto}\nonumber \\ 
    & \quad = \mathcal{L}_{pBCE}(X_s,\hat{Y}_s;P_s^{linear})+ \mathcal{L}_{proto}(X_s,\hat{Y}_s;P_s^{lproto}) \nonumber \\
    & \quad + \mathcal{L}_{proto}(X_s,\hat{Y}_s;P_s^{ulproto}).
    \label{eq:loss_strongview}
\end{align}

Equation~\ref{eq:loss_strongview} applies consistency regularization to the three classifiers of the student model. By constraining the consistency of the labeled and unlabeled prototype classifiers, we gradually align the feature distribution of labeled pixels with that of unlabeled pixels. By encouraging the consistency of linear and prototype classifiers, we improve the intra-class compactness and inter-class separability. Such a regularization scheme implicitly modulates the feature representation: features from the same category should be close to both the corresponding prototype of the labeled pixels and the prototype of the unlabeled pixels, while being far away from prototypes of other classes. 

The implementation of our prototype classifier follows the work of ~\citep{zhou2022rethinking}. To address intra-class heterogeneity, $K$ prototypes $\{{\rm{p}_c^k}\}_{k=1}^K$ are generated for each class $c$ using online clustering. The prototypes are updated at each iteration with a momentum coefficient $\mu=0.999$ as $\rm{p}_c^k=\mu \rm{p}_c^k+(1-\mu) \widetilde{\rm{p}_c^k}$, where $\widetilde{\rm{p}_c^k}$ is the prototypes computed at the current iteration. By calculating the distance between each pixel and the prototypes, the segmentation prediction $p(c|i)$ of the prototype classifier is obtained. This process can be defined as
\begin{equation}
    p(c|i)=\frac{exp(-min\{{<i,{\rm{p}}_c^k>}\}_{k=1}^K)}{\sum_{c^{\prime}=0}^{C}{exp(-min\{{<i,{\rm{p}}_{c^{\prime}}^k>}\}_{k=1}^K)}},
    \label{eq:prototype segmentation}
\end{equation}
where $i$ denotes the $l_2$ normalized embedding of pixel $i$, and the dimension of $i$ is 1$\times$64. The distance measurement $<\cdot,\cdot>$ denotes the negative cosine similarity, and $<i,{\rm{p}_c^k}>=-i^T {\rm{p}_c^k}$. $min\{{<i,\rm{p}_c^k>}\}_{k=1}^K$ denotes the distance from the embedding of pixel $i$ to its nearest prototype. The loss used to optimize the prototype classifier is
\begin{align}
    & \mathcal{L}_{proto}(X_s,\hat{Y}_s) = \mathcal{L}_{CE}+\lambda_1 \mathcal{L}_{PPD}+\lambda_2 \mathcal{L}_{PPC} \nonumber \\
    & \quad = \sum_{c=0}^C \sum_{k=1}^K \sum_{i \in X_s} (-\hat{y}_c\log p(c|i)+\lambda_1(1-i^T {\rm{p}}_c^k)^2 -  \nonumber \\  
    & \quad \lambda_2 \log({\frac{exp(i^T {\rm{p}}_c^k / \alpha)}{exp(i^T {\rm{p}}_c^k / \alpha) + \sum_{{\rm{p}}^- \in {\mathcal{P}}^-} exp({ i^T {\rm{p}}^-/\alpha})}})),
    \label{eq:proto_classifier loss}
\end{align}
where $\hat{y}_c$ is the masked pseudo-label corresponding to category $c$. $\mathcal{L}_{CE}$ is the cross-entropy loss that forces the embedding of pixel $i$ to be proximate to the prototypes of its corresponding class $c$ while remaining far away from prototypes of other classes. Only focusing on the training objective $\mathcal{L}_{CE}$ is insufficient. The optimization of $\mathcal{L}_{CE}$ solely improves the inter-class and intra-class relative relationship. However, when the intra-class distance is smaller than other inter-class distances, the penalty from $\mathcal{L}_{CE}$ becomes insignificant, even if the intra-class distance is substantial. To tackle this problem, $\mathcal{L}_{ppd}$ minimizes the distance between each embedded pixel $i$ and its assigned prototype ${\rm{p}}_c^k$ directly, reducing the intra-class variation. Additionally, $\mathcal{L}_{CE}$ does not consider within-class pixel-prototype relations. $\mathcal{L}_{ppc}$ enforces each pixel embedding $i$ to be similar to its assigned positive prototype ${\rm{p}}_c^k$ and dissimilar to other $CK-1$ irrelevant negative prototypes ${\mathcal{P}}^-$ which denotes ${\{{\rm{p}}_c^{k^{\prime}}\}_{c=0,k^{\prime}=1}^{C,K}}/{\rm{p}}_c^k$. The parameters $\alpha$, $\lambda_1$, and $\lambda_2$ are set to 0.1, 0.001, and 0.01, respectively.

The training procedure of our method is shown in Algorithm ~\ref{alg:ours}. Since the prototype-based classifier relies on non-learnable prototypes for classification, the quality of the initial prototypes is critical. Therefore, we first train a linear classifier in the first stage. In the second stage, we use the model trained in the first stage for initialization and then train the linear and two prototype classifiers together. One of the key strengths of our proposed LTUDA framework is its high efficiency. Firstly, the CDA module performs data mixing at the input level without introducing additional parameters. Secondly, in the PDA module, both prototype classifiers rely on non-parametric prototypes for prediction without adding too much extra overhead. Furthermore, regularization is only performed during training, and only one classifier is required to make predictions during inference, so the inference time does not increase.
\begin{table}
\tiny
\renewcommand\arraystretch{1.2}
\setlength{\belowcaptionskip}{4pt}
\caption{A brief description of the large-scale partially-labeled dataset.}
\label{tab:dataset_details}
\centering
\resizebox{\linewidth}{!}{
\begin{tabular}{l|c|c|c}
\hline
\multirow{2}{*}{Dataset} & \multirow{2}{*}{Annotations} & \multicolumn{2}{c}{ Number} \\ \cline{3-4} 
 & & Training & Testing  \\ \hline
LiTS & \#1 Liver & 84 & 22 \\ \hline
MSD-Spleen & \#2 Spleen & 33 & 8 \\ \hline
KiTS & \#3 Kidney & 164 & 40 \\ \hline
NIH82 & \#4 Pancreas & 66 & 16 \\ \hline
BTCV & All Organs & - & 30 \\ \hline
AbdomenCT-1K & All Organs & - & 50 \\ \hline 
Total & -  & 347 & 166  \\ 
\hline 
\end{tabular}}
\end{table}

\begin{table*}[ht]
\scriptsize
\renewcommand\arraystretch{1.2}
\setlength{\belowcaptionskip}{4pt}
\caption{Quantitative results of partially-supervised multi-organ segmentation on a toy dataset.}
\label{tab:SOTA_toy}
\centering
\resizebox{\linewidth}{!}{
\begin{tabular}
{l | c c c c c | c c c c c}
\shline
\multirow{2}{*}{Method}&
\multicolumn{5}{c|}{Dice(\%) $\uparrow$} & \multicolumn{5}{c}{HD $\downarrow$}
\\ \cline{2-6} \cline{7-11} & liver  & spleen  & kidney & pancreas  & avg & liver  & spleen  & kidney & pancreas  & avg  \\ 
\hline
Fully-supervised &98.35 &97.66 &94.81 &77.50 &92.08 &\textbf{9.29} &3.61 &13.66 &12.99 &9.89 \\ 
\hline
Multi-Nets &97.35 &95.35 &94.35 &63.88 &87.73 &15.21 &5.46 &13.32 &16.90 &12.72 \\ 
Med3d~\citep{chen2019med3d} &97.23 &95.57 &93.57 &60.76 &86.78 &13.93 &5.43 &\textbf{12.41} &27.12 &14.72 \\ 
Dodnet~\citep{zhang2021dodnet} &97.71 &96.83 &93.47 &63.40 &87.85 &12.11 &4.05 &14.27 &20.85 &12.82  \\
PIPO-FAN~\citep{fang2020multi} &96.01 &96.90 &93.08 &62.05 &87.01 &15.61 &4.38 &18.23 &26.37 &16.15 \\ 
Mloss and Eloss~\citep{shi2021marginal} &97.75 &96.54 &93.70 &65.77 &88.44 &12.08 &4.62 &14.33 &23.59 &13.66 \\ 
Self-training~\citep{lee2013pseudo} &97.64 &96.59 &93.60 &69.64 &89.37 &13.13 &4.02 &17.21 &20.16 &13.63 \\ 
Co-training~\citep{huang2020multi} &97.65 &96.57 &94.25 &70.49 &89.74 &12.54 &4.00 &14.94 &17.73 &12.30 \\ 
Ms-kd~\citep{feng2021ms} &97.69 &96.26 &94.43 &70.64 &89.75 &13.28 &4.58 &13.33 &16.25 &11.86 \\ 
CPS~\citep{chen2021semi} &97.68 &96.81 &94.51 &68.83 &89.45 &12.21 &4.18 &12.61 &19.05 &12.16 \\ 
DMPLS~\citep{luo2022scribble} &97.66 &97.00 &94.01 &67.76 &89.11 &12.67 &3.74 &13.46 &18.46 &12.08 \\ 
\hline
LTUDA (Ours) &\textbf{98.39} &\textbf{97.70} &\textbf{95.23} &\textbf{80.43} &\textbf{92.94} &9.49 &\textbf{3.39} &12.63 &\textbf{12.04} &\textbf{9.39} \\ 
\shline
\end{tabular}
}
\end{table*}

\begin{table*}[ht]
\scriptsize
\renewcommand\arraystretch{1.2}
\setlength{\belowcaptionskip}{4pt}
\caption{Quantitative results of partially-supervised multi-organ segmentation on four partially labeled datasets (LSPL dataset).}
\label{tab:SOTA}
\centering
\resizebox{\linewidth}{!}{
\begin{tabular}
{l| ccccc|ccccc}
\shline
\multirow{2}{*}{Method}&
\multicolumn{5}{c|}{Dice(\%) $\uparrow$} & \multicolumn{5}{c}{HD $\downarrow$} 
\\ \cline{2-6} \cline{7-11}  & liver  & spleen  & kidney & pancreas  & avg & liver  & spleen  & kidney & pancreas  & avg  \\ 
\hline
Multi-Nets &94.17 &86.39 &94.13 &74.17 &87.22 &15.16 &11.39 &12.52 &15.23 &13.58 \\ 
Med3d~\citep{chen2019med3d} &94.74 &89.75 &94.60 &75.13 &88.56 &14.74 &9.39 &10.87 &16.00 &12.75  \\ 
Dodnet~\citep{zhang2021dodnet} &94.48 &89.06 &94.71 &75.04 &88.32 &14.56 &9.51 &10.68 &14.11 &12.22   \\ 
PIPO-FAN~\citep{fang2020multi} &94.47 &88.24 &94.02 &67.64 &86.09 &16.28 &10.47 &12.67 &17.46 &14.22 \\ 
Mloss and Eloss~\citep{shi2021marginal} &94.98 &91.73 &94.55 &74.61 &88.97 &14.59 &7.57 &10.84 &14.27 &11.82  \\ 
Self-training~\citep{lee2013pseudo} &94.58 &87.02 &94.37 &76.03 &88.00 &13.97 &10.97 &10.87 &16.54 &13.09  \\ 
Co-training~\citep{huang2020multi} &95.29 &87.78 &94.30 &76.97 &88.59 &13.56 &10.50 &11.40 &13.95 &12.35  \\ 
Ms-kd~\citep{feng2021ms} &94.63 &86.50 &94.20 &74.88 &87.55 &14.06 &11.36 &11.59 &14.05 &12.76   \\ 
CPS~\citep{chen2021semi} &\textbf{95.65} &92.08 &94.77 &73.46 &88.99 &\textbf{12.90} &6.78 &\textbf{10.46} &16.49 &11.66  \\ 
DMPLS~\citep{luo2022scribble} &95.38 &91.81 &94.64 &71.26 &88.27 &13.32 &7.00 &10.79 &18.32 &12.36  \\ 
\hline
LTUDA (Ours) &95.36 &\textbf{92.88} &\textbf{94.79} &\textbf{81.27} &\textbf{91.08} &14.30 &\textbf{6.37} &\textbf{10.46} &\textbf{11.40} &\textbf{10.63}  \\ 
\shline
\end{tabular}
}
\end{table*}

\section{Experiments}
\subsection{Datasets}
\myparagraph{Toy dataset.} In our experiments, to evaluate the multi-organ segmentation performance and measure the performance gap between partially-supervised and fully-supervised methods, we sample a partially-labeled toy dataset from the AbdomenCT-1K dataset~\citep{ma2021abdomenct}. AbdomenCT-1K is a large and diverse abdominal dataset, which includes a subset of 50 cases (from Nanjing University) that have annotations for 13 organs. We use these 50 cases as a toy dataset, and 40 of the cases are randomly selected as a training set. The training set is divided into four partially labeled datasets, each providing annotations for the liver, spleen, kidney, and pancreas, respectively. The remaining 10 cases are used as the test set, with annotations of these four organs.

\myparagraph{LSPL dataset.} We construct a large-scale partially-labeled dataset (called the LSPL dataset) using four abdominal single-organ benchmark datasets: LiTS~\citep{bilic2023liver}, MSD-Spleen~\citep{simpson2019large}, KiTS~\citep{heller2019kits19}, and NIH82~\citep{roth2015deeporgan}. These four single-organ datasets provide annotations for the liver, spleen, kidney, and pancreas, respectively. A total of 433 abdominal CT scans are collected from these partially labeled datasets, of which 347 are used for training and 86 for testing. Table~\ref{tab:dataset_details} provides detailed information. Furthermore, we evaluate the performance of the segmentation model on two multi-organ datasets, BTCV (Beyond The Cranial Vault) ~\citep{landman2017multi} and the AbdomenCT-1K dataset~\citep{ma2021abdomenct}. BTCV contains 47 CT scans with annotations for 13 organs, but only 30 of them have annotations for the kidney. The Abdomen1K dataset contains a total of 1112 CT scans and comprises five public datasets (LiTS, KiTS, MSD-Spleen, MSD-Pancreas~\citep{simpson2019large}, and NIH82) and a new dataset from Nanjing University. Since our training data includes LiTS, KiTS, MSD-Spleen, and NIH82, here, we use 50 cases from Nanjing University for evaluation (same as the toy dataset). 

The modality of the images is abdominal CT scans, with the assumption that the position/posture remains consistent. Images in different datasets vary in spatial resolutions and image sizes. For data preprocessing, we truncate the Hounsfield unit (HU) values to the range of $[-325, 325]$ to better contrast abdominal organs, and linearly normalize the data to the range of $[-1, 1]$. Notably, we maintain the diversity of training datasets collected from different institutions, including a variety of slice spacings and pathologies. We slice 3D cases along the z-axis and remove irrelevant non-abdominal regions, and finally resize the axial slices into $256\times256$ pixels as input to the network.

\subsection{Implementation details.}
We employ a 2D U-Net~\citep{ronneberger2015u} as the segmentation backbone. All experiments are implemented with the Pytorch library and are performed with a batch size of 4 on one NVIDIA 3090 GPU with 24GB memory. The parameters $\tau$, $M$, and $K$ are set as 0.5, 2, and 5, respectively. The network is optimized using the SGD optimizer with an initial learning rate of $1e-3$, and the learning rate decays according to the policy ${\rm lr}={\rm lr}_{init}\times(1-t/T)^{0.9}$, where $T$ is the maximum epochs. We quantitatively measure the performance of different methods by the Dice similarity coefficient (Dice, higher is better) and Hausdorff distance (HD, lower is better).
\begin{table*}[ht]
\scriptsize
\renewcommand\arraystretch{1.2}
\setlength{\belowcaptionskip}{4pt}
\caption{Generalization performance of different methods on two multi-organ datasets.}
\label{tab:SOTA_multi}
\centering
\resizebox{\linewidth}{!}{
\begin{tabular}
{l|ccccc|ccccc}
\shline
\multirow{2}{*}{Method}
&\multicolumn{5}{c|}{Dice(\%) $\uparrow$} & \multicolumn{5}{c}{HD $\downarrow$} \\ \cline{2-6} \cline{7-11}
& liver  & spleen  & kidney & pancreas  & avg & liver  & spleen  & kidney & pancreas  & avg  \\ 
\hline
\multicolumn{11}{c}{BTCV} \\ 
\hline
Multi-Nets &95.14 &89.33 &87.07 &70.76 &85.58 &17.08 &8.89 &19.18 &18.04 &15.80 \\ 
Med3d~\citep{chen2019med3d} &95.04 &91.50 &87.61 &71.82 &86.49 &15.94 &7.05 &14.64 &20.21 &14.46 \\
Dodnet~\citep{zhang2021dodnet} &95.54 &91.56 &87.44 &71.94 &86.62 &16.19 &9.00 &14.70 &19.32 &14.80  \\
PIPO-FAN~\citep{fang2020multi} &95.21 &92.03 &87.06 &61.15 &83.86 &15.92 &7.96 &16.15 &21.87 &15.48 \\ 
Mloss and Eloss~\citep{shi2021marginal} &95.46 &93.17 &88.17 &70.46 &86.81 &16.27 &7.22 &\textbf{14.55} &20.33 &14.59 \\ 
Self-training~\citep{lee2013pseudo} &95.53 &91.03 &87.91 &71.38 &86.46 &16.04 &9.69 &15.57 &18.75 &15.01 \\ 
Co-training~\citep{huang2020multi} &95.60 &91.86 &88.12 &72.75 &87.08 &16.18 &8.63 &15.69 &18.64 &14.79 \\ 
Ms-kd~\citep{feng2021ms} &95.24 &90.47 &87.81 &71.79 &86.33 &16.72 &8.41 &15.75 &18.82 &14.92 \\  
CPS~\citep{chen2021semi} &95.91 &94.04 &87.82 &68.44 &86.55 &15.54 &5.94 &15.58 &21.87 &14.74 \\ 
DMPLS~\citep{luo2022scribble} &\textbf{96.04} &93.45 &88.07 &66.72 &86.07 &\textbf{14.88} &6.55 &15.02 &22.40 &14.71 \\ 
\hline
LTUDA (Ours) &95.98 &\textbf{95.18} &\textbf{88.69} &\textbf{80.06} &\textbf{89.98} &15.34 &\textbf{4.85} &16.11 &\textbf{14.14} &\textbf{12.61} \\ 
\hline 
\multicolumn{11}{c}{AbdomenCT-1K} \\ 
\hline 
Multi-Nets &97.91 &95.57 &92.88 &65.94 &88.08 &9.63 &4.50 &14.79 &17.68 &11.65 \\ 
Med3d~\citep{chen2019med3d} &97.96 &96.14 &94.73 &67.07 &88.98 &9.82 &4.02 &12.06 &17.33 &10.81 \\ 
Dodnet~\citep{zhang2021dodnet} &97.86 &95.63 &93.98 &65.41 &88.22 &9.58 &3.98 &13.00 &17.66 &11.06  \\ 
PIPO-FAN~\citep{fang2020multi} &97.86 &95.82 &94.14 &56.84 &86.16 &9.85 &4.16 &11.69 &21.49 &11.80 \\ 
Mloss and Eloss~\citep{shi2021marginal} &97.98 &96.13 &94.48 &65.84 &88.61 &9.78 &3.66 &11.91 &16.89 &10.56 \\ 
Self-training~\citep{lee2013pseudo} &97.94 &95.80 &94.30 &68.07 &89.03 &9.33 &4.23 &11.67 &17.20 &10.61 \\ 
Co-training~\citep{huang2020multi} &97.86 &96.39 &94.02 &70.34 &89.65 &\textbf{9.27} &3.69 &12.85 &16.30 &10.53 \\ 
Ms-kd~\citep{feng2021ms} &97.98 &95.79 &93.70 &64.84 &88.08 &9.55 &4.26 &13.53 &17.68 &11.25 \\ 
CPS~\citep{chen2021semi} &\textbf{98.08} &96.59 &\textbf{95.08} &73.88 &90.91 &8.94 &\textbf{3.41} &\textbf{11.08} &18.08 &10.38 \\ 
DMPLS~\citep{luo2022scribble} &97.99 &96.16 &94.24 &72.22 &90.15 &9.29 &3.56 &12.34 &17.56 &10.69 \\ 
\hline
LTUDA (Ours) &97.54 &\textbf{96.91} &94.82 &\textbf{78.42} &\textbf{91.92} &11.66 &3.50 &12.04 &\textbf{12.77} &\textbf{9.99} \\ 
\shline
\end{tabular}}
\end{table*}
\begin{table*}
\Large
\renewcommand\arraystretch{1.2}
\setlength{\belowcaptionskip}{4pt}
\caption{Ablation study of key components on the toy dataset. CDA denotes cross-set data augmentation, and PDA denotes prototype-based distribution alignment. Green numbers indicate the performance improvement over the baseline. }
\label{tab:ablation_CR}
\centering
\resizebox{\linewidth}{!}{
\begin{tabular}{ccc| ccccc|ccccc}
\shline
\multicolumn{3}{c|}{Method}&
\multicolumn{5}{c|}{Dice(\%) $\uparrow$} & \multicolumn{5}{c}{HD $\downarrow$}
\\ \cline{1-3} \cline{4-8} \cline{9-13}  baseline &CDA &PDA  & liver  & spleen  & kidney & pancreas  & avg & liver  & spleen  & kidney & pancreas  & avg  \\ 
\hline
\checkmark & & &97.40 &96.42 &94.44 &63.04 &87.82 &12.70 &4.65 &14.20 &17.49 &12.26 \\ 
\checkmark &\checkmark & &\reshl{98.29}{0.89} &\reshl{\textbf{97.71}}{1.29} &\reshl{94.57}{0.13} &\reshl{77.63}{14.59} &\reshl{92.05}{4.23} &\reshld{9.66}{3.04} &\reshld{\textbf{2.87}}{1.78} &\reshl{14.47}{0.27} &\reshld{15.18}{2.31} &\reshld{10.55}{1.71}  \\ 
\checkmark &\checkmark &\checkmark &\reshl{\textbf{98.39}}{0.99} &\reshl{97.70}{1.28} &\reshl{\textbf{95.23}}{0.79} &\reshl{\textbf{80.43}}{17.39} &\reshl{\textbf{92.94}}{5.12} &\reshld{\textbf{9.49}}{3.21} &\reshld{3.39}{1.26} &\reshld{\textbf{12.63}}{1.57} &\reshld{\textbf{12.04}}{5.45} &\reshld{\textbf{9.39}}{2.87} \\  
\shline
\end{tabular}}
\end{table*}


\subsection{Comparison with the state-of-the-art}

We compare the proposed method with other state-of-the-art methods: (1) Multi-Nets, which trains four separate networks, each focusing on segmenting a specific organ; (2) two condition-based methods, namely  Med3d~\citep{chen2019med3d} and DoDNet~\citep{zhang2021dodnet}; (3) two unified training methods, namely PIPO-FAN~\citep{fang2020multi}, Mloss and Eloss~\citep{shi2021marginal}; (4) several two-stage pseudo-labeling methods, which require pre-training multiple single-organ segmentation models to generate pseudo-labels for unlabeled pixels, the self-training~\citep{lee2013pseudo}, co-training~\cite{huang2020multi} and Ms-kd~\cite{feng2021ms} methods; (5) several one-stage pseudo-labeling methods, which generate pseudo-labels for unlabeled categories online, namely, CPS~\citep{chen2021semi} and DMPLS~\citep{luo2022scribble}. Among the two-stage pseudo-labeling methods, the co-training
method jointly trains a pair of weight-averaged multi-organ models to tolerate noise in the pseudo-labels, while Ms-kd introduces knowledge distillation between the single-organ teacher models and the multi-organ student model. The one-stage pseudo-labeling methods CPS and DMPLS train a dual-branch network. CPS enforces cross-supervision between the two branches, while DMPLS leverages the dynamic ensemble results of both branches as supervision. To ensure a fair comparison, we use the same backbone as our method for all re-implemented methods. All compared methods use weak augmentation. It is noteworthy that, in their original papers, methods PIPO-FAN~\citep{fang2020multi},  Mloss and Eloss~\citep{shi2021marginal} require a portion of fully labeled data to be trained together with partially labeled data. In comparison, in our implementation, only partially labeled data are used as the training set.

\begin{figure}[t]
    \centering
    \setlength{\abovecaptionskip}{-0.25cm}
    \includegraphics[width=0.48\textwidth,
    height=0.7\textwidth]{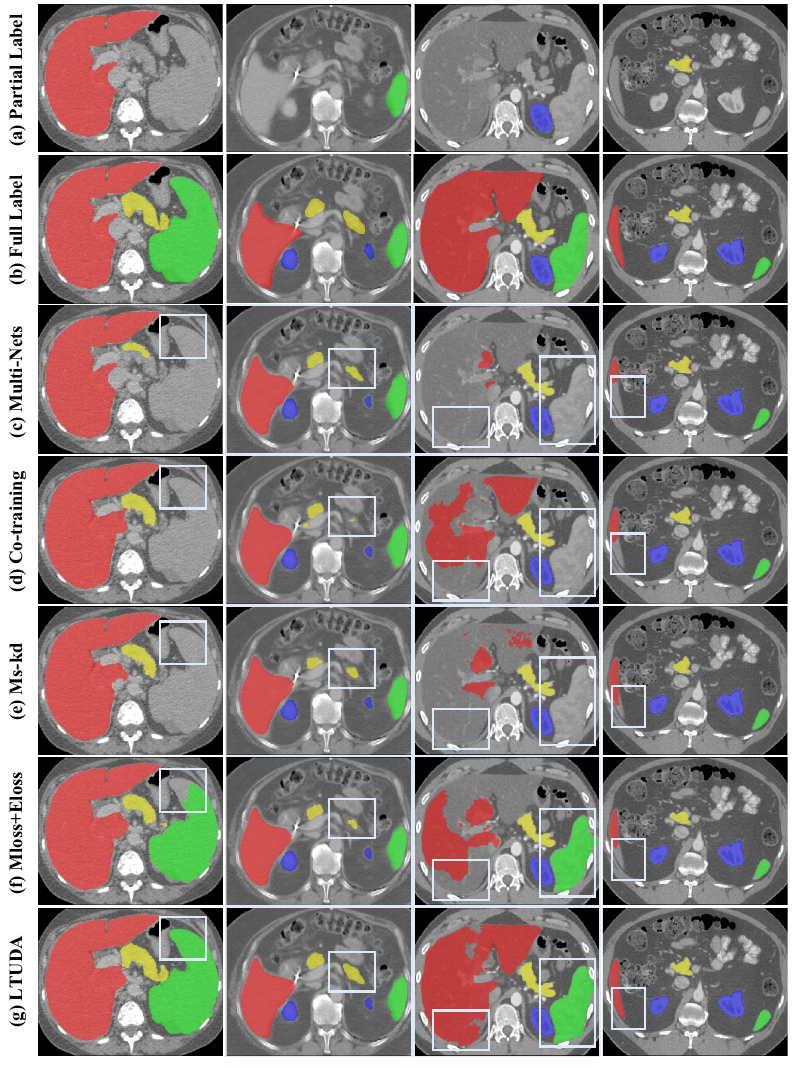}
	\caption{Visualizations of LSPL. Examples are from the LiTS, MSD-Spleen, KiTS, and NIH82 datasets, respectively, arranged from left to right. (a) Single-organ annotations originally provided by the four benchmark datasets. (b) Full annotations of four organs. (c) to (g) are the segmentation results of different methods. The white frame highlights the better predictions of our method.}
	\label{vis}
\end{figure}

\myparagraph{Results on the toy dataset.} On the toy dataset, we report the performance of fully-supervised learning as an upper bound. The quantitative performance is presented in Table~\ref{tab:SOTA_toy}, where the following results  can be observed: (1) Most methods (except Med3d and PIPO-FAN) achieve better performance than Multi-Nets, indicating that joint training of partially labeled datasets can improve model performance. (2) The pseudo-labeling methods (self-training, co-training, Ms-kd, CPS, and DMPLS) outperform the other methods. This demonstrates the advantage of fully exploiting large amounts of unlabeled pixels in improving performance. (3) Our proposed method achieves the highest overall performance with an average Dice of 92.94\% and an average HD of 9.39, which demonstrates its effectiveness. Furthermore, our method obtains the best performance on Dice for all four organs, significantly improving the pancreas category with blurred boundaries. (4) Our method achieves superior performance over fully-supervised learning. This improvement is attributable to two factors. First, the CDA module expands the training data by incorporating unlabeled pixels. Second, the PDA module improves both intra-class consistency and intra-class separation.

\myparagraph{Results on the LSPL dataset.} The quantitative performance on the LSPL dataset is shown in Table~\ref{tab:SOTA}. Notably, here we utilize the re-annotations provided by the AbdomenCT-1K dataset to evaluate the four-organ segmentation predictions. The AbdomenCT-1K dataset had re-annotated LiTS, MSD-Spleen, KiTS, and NIH82, retaining their original single-organ annotations and adding additional annotations for the remaining three categories. This enables us to exploit these re-annotations to evaluate the four-organ predictions of different models.

It can be seen that our method achieves the best overall segmentation accuracy, with an average Dice of 91.08\% and an average HD of 10.63. Our method significantly outperforms state-of-the-art partially-supervised segmentation methods, especially for the pancreas and spleen categories. The substantial improvement in the segmentation performance of the pancreas and spleen can be attributed to two factors. Firstly, the class imbalance issue is evident, as demonstrated in Table~\ref{tab:dataset_details}, where the number of training samples for the spleen and pancreas is fewer than that for the liver and kidney. Moreover, the pancreas is relatively small compared to the other organs, resulting in fewer labeled pixels for the pancreas compared to other categories. Secondly, different organ segmentation tasks possess varying difficulty levels, with the pancreas particularly challenging. The pancreas is characterized by its relatively small size, low contrast with surrounding organs, and indistinct boundaries. Even under fully-supervised segmentation, the segmentation accuracy of the pancreas remains lower than that of other organs. The aforementioned factors contribute to more substantial distribution mismatch issues and spleen categories in contrast to the liver and kidney categories.

\begin{table*}[ht]
\large
\renewcommand\arraystretch{1.2}
\setlength{\belowcaptionskip}{4pt}
\caption{Ablation study of key components on LSPL dataset. CDA denotes cross-set data augmentation, and PDA denotes prototype-based distribution alignment. Green numbers indicate the performance improvement over the baseline.}
\label{tab:ablation_LSPL}
\centering
\resizebox{\linewidth}{!}{
\begin{tabular}{ccc| ccccc|ccccc}
\shline
\multicolumn{3}{c|}{Method}&
\multicolumn{5}{c|}{Dice(\%) $\uparrow$} & \multicolumn{5}{c}{HD $\downarrow$}
\\ \cline{1-3} \cline{4-8} \cline{9-13}  baseline &CDA &PDA  & liver  & spleen  & kidney & pancreas  & avg & liver  & spleen  & kidney & pancreas  & avg  \\
\hline
\multicolumn{11}{c}{LSPL} \\ 
\hline
\checkmark & & &93.67 &88.44 &94.62 &75.53 &88.07 &15.59 &10.19 &10.96 &16.04 &13.19 \\ 
\checkmark &\checkmark & &\reshl{\textbf{95.37}}{1.7} &\reshl{92.41}{3.97} &\reshl{\textbf{94.80}}{0.18} &\reshl{80.36}{4.83} &\reshl{90.74}{2.67} &\reshld{\textbf{13.95}}{1.64} &\reshld{6.53}{3.66} &\reshld{\textbf{10.45}}{0.51} &\reshld{12.01}{4.03} &\reshld{10.74}{2.45}  \\ 
\checkmark &\checkmark &\checkmark &\reshl{95.36}{1.69} &\reshl{\textbf{92.88}}{4.44} &\reshl{94.79}{0.17} &\reshl{\textbf{81.27}}{5.74} &\reshl{\textbf{91.08}}{3.01} &\reshld{14.30}{1.29} &\reshld{\textbf{6.37}}{3.82} &\reshld{10.46}{0.5} &\reshld{\textbf{11.40}}{4.64} &\reshld{\textbf{10.63}}{2.56} \\  
\hline 
\multicolumn{11}{c}{BTCV} \\ 
\hline 
\checkmark & & &95.05 &92.49 &87.68 &70.22 &86.36 &17.54 &7.99 &15.67 &19.53 &15.18 \\ 
\checkmark &\checkmark & & \reshl{95.94}{0.89} &\reshl{94.83}{2.34} &\reshl{88.59}{0.91} &\reshl{78.53}{8.31} &\reshl{89.47}{3.11} &\reshld{15.52}{2.02} &\reshld{4.99}{3.00} &\reshld{\textbf{14.01}}{1.66} &\reshld{14.82}{4.71} &\reshld{12.34}{2.84} \\ 
\checkmark &\checkmark &\checkmark &\reshl{\textbf{95.98}}{0.93} &\reshl{\textbf{95.18}}{2.69} &\reshl{\textbf{88.69}}{1.01} &\reshl{\textbf{80.06}}{9.84} &\reshl{\textbf{89.98}}{3.62} &\reshld{\textbf{15.34}}{2.2} &\reshld{\textbf{4.85}}{3.14} &\reshl{16.11}{0.44} &\reshld{\textbf{14.14}}{5.39} &\reshld{\textbf{12.61}}{2.57}  \\  
\hline
\multicolumn{11}{c}{AbdomenCT-1K} \\ 
\hline
\checkmark & & &97.56 &95.99 &94.11 &67.87 &88.88 &10.95 &4.33 &12.19 &17.87 &11.34 \\ 
\checkmark &\checkmark & &\reshl{\textbf{97.75}}{0.19} &\reshl{\textbf{97.11}}{1.12} &\reshl{94.28}{0.17} &\reshl{74.38}{6.51} &\reshl{90.88}{2.00} &\reshld{\textbf{10.36}}{0.59} &\reshld{\textbf{3.12}}{1.21} &\reshld{\textbf{11.80}}{0.39} &\reshld{14.19}{3.68} &\reshld{\textbf{9.87}}{1.47}  \\ 
\checkmark &\checkmark &\checkmark &\reshld{97.54}{0.02} &\reshl{96.91}{0.92} &\reshl{\textbf{94.82}}{0.71} &\reshl{\textbf{78.42}}{10.55} &\reshl{\textbf{91.92}}{3.04} &\reshl{11.66}{0.71} &\reshld{3.50}{0.83} &\reshld{12.04}{0.15} &\reshld{\textbf{12.77}}{5.1} &\reshld{9.99}{1.35} \\
\shline
\end{tabular}}
\end{table*}

\myparagraph{Visualized results.} We can observe significant improvements in the multi-organ segmentation results, as shown in Fig.~\ref{vis}. Compared to the toy dataset, the distribution mismatch is more pronounced in the LSPL dataset due to domain gaps between data from different institutions. The Multi-Nets models trained individually on single-organ datasets have poor generalization performance, resulting in a large number of false negative predictions. For example, although the liver segmentation model trained on LiTS can perform well on the test set of LiTS, it generalizes poorly on images from MSD-Spleen, KiTS, and NIH82. Pseudo-labeling methods (\eg, co-training and Ms-kd) train student models based on pseudo-labels generated by Multi-Nets. Biased pseudo-labels lead to errors accumulating such that the student model predicts similar errors. The Mloss and Eloss method treats unlabeled pixels as background, which causes confusion between the true background and unlabeled foreground, resulting in under-segmented predictions. Our method achieves the overall best performance on multi-organ segmentation. 

To further demonstrate the effectiveness of our proposed method, we also evaluate its generalization ability on two external multi-organ datasets ( BTCV and the AbdomenCT-1K dataset). Note that the model trained solely from the LSPL dataset, which has not seen images from the BTCV and AbdomemCT-1K datasets. The quantitative performance is shown in Table~\ref{tab:SOTA_multi}. On both datasets, our method achieves state-of-the-art performance, ranking among the top for almost all evaluation metrics in every category, prominently outperforming the other methods. These results indicate that our method has excellent generalization capability.

\subsection{Ablation studies}
\begin{figure}[tbp]
    \centering
    \setlength{\abovecaptionskip}{-0.5cm}
    \includegraphics[width=0.48\textwidth]{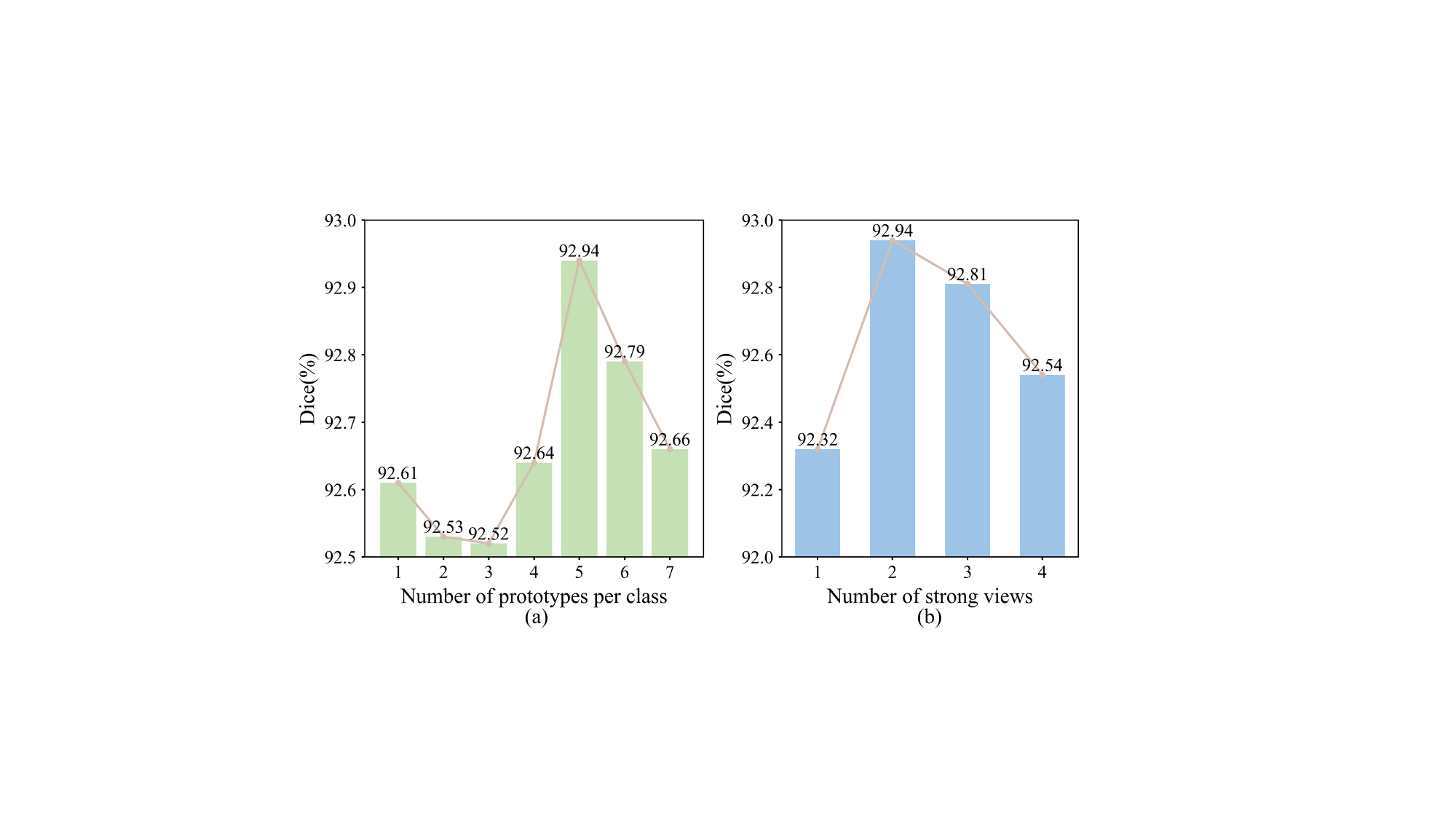}
	\caption{ Ablation results. (a) The number of prototypes per class. (b) The number of strong views. }
	\label{ablation_fig1}
\end{figure}

We further conduct ablation experiments on the toy dataset to verify the effectiveness of the proposed method.

\begin{table*}
\scriptsize
\renewcommand\arraystretch{1.2}
\setlength{\belowcaptionskip}{4pt}
\caption{Ablation results of different data augmentation methods. }
\label{tab:ablation_DA}
\centering
\resizebox{\linewidth}{!}{
\begin{tabular}{cc| ccccc|ccccc}
\shline
\multicolumn{2}{c|}{} & \multicolumn{5}{c|}{Dice(\%) $\uparrow$} & \multicolumn{5}{c}{HD $\downarrow$}                                   \\ \cline{3-7} \cline{8-12}  
\multicolumn{2}{c|}{\multirow{-2}{*}{Method}}                                      & \multicolumn{1}{c}{liver} & \multicolumn{1}{c}{spleen} & \multicolumn{1}{c}{kidney} & \multicolumn{1}{c}{pancreas} & avg            & \multicolumn{1}{c}{liver} & \multicolumn{1}{c}{spleen} & \multicolumn{1}{c}{kidney} & \multicolumn{1}{c}{pancreas} & avg            \\ 
\hline 
\multicolumn{2}{c|}{Baseline}                                                      & \multicolumn{1}{c}{97.40} & \multicolumn{1}{c}{96.42}  & \multicolumn{1}{c}{94.44}  & \multicolumn{1}{c}{63.04}    & 87.82          & \multicolumn{1}{c}{12.70} & \multicolumn{1}{c}{4.65}   & \multicolumn{1}{c}{14.20}  & \multicolumn{1}{c}{17.49}    & 12.26          \\ 
\hline 
\multicolumn{1}{c}{}                                               & Cutout       & \multicolumn{1}{c}{98.08} & \multicolumn{1}{c}{96.98}  & \multicolumn{1}{c}{94.88}  & \multicolumn{1}{c}{69.82}    & 89.94          & \multicolumn{1}{c}{10.94} & \multicolumn{1}{c}{3.89}   & \multicolumn{1}{c}{\textbf{12.43}}  & \multicolumn{1}{c}{20.94}    & 12.05          \\  
\multicolumn{1}{c}{\multirow{-2}{*}{Intra-set  }} & Color jitter & \multicolumn{1}{c}{98.02} & \multicolumn{1}{c}{96.69}  & \multicolumn{1}{c}{93.65}  & \multicolumn{1}{c}{67.95}    & 89.08          & \multicolumn{1}{c}{11.82} & \multicolumn{1}{c}{4.12}   & \multicolumn{1}{c}{12.66}  & \multicolumn{1}{c}{17.78}    & 11.60          \\ 
\hline

\multicolumn{1}{c}{}                                               & Cutmix       & \multicolumn{1}{c}{98.29} & \multicolumn{1}{c}{\textbf{97.71}}  & \multicolumn{1}{c}{94.57}  & \multicolumn{1}{c}{77.63}    & 92.05          & \multicolumn{1}{c}{9.66} & \multicolumn{1}{c}{\textbf{2.87}}   & \multicolumn{1}{c}{14.47}  & \multicolumn{1}{c}{15.18}    & 10.55          \\ 
\multicolumn{1}{c}{}                                               & Cowmix       & \multicolumn{1}{c}{\textbf{98.31}} & \multicolumn{1}{c}{97.51}  & \multicolumn{1}{c}{94.06}  & \multicolumn{1}{c}{\textbf{78.03}}    & 91.98          & \multicolumn{1}{c}{\textbf{9.65}} & \multicolumn{1}{c}{3.26}   & \multicolumn{1}{c}{15.56}  & \multicolumn{1}{c}{\textbf{13.47}}    & \textbf{10.48}          \\ 

\multicolumn{1}{c}{\multirow{-2}{*}{Cross-set  }} & Carvemix & \multicolumn{1}{c}{98.18} & \multicolumn{1}{c}{\textbf{97.71}}  & \multicolumn{1}{c}{\textbf{94.92}}  & \multicolumn{1}{c}{77.42}    & \textbf{92.06}          & \multicolumn{1}{c}{10.44} & \multicolumn{1}{c}{\textbf{2.87}}   & \multicolumn{1}{c}{14.12}  & \multicolumn{1}{c}{14.68}    & 10.53          \\ 
\shline
\end{tabular}}
\end{table*}


\myparagraph{Effectiveness of core modules.} 
To analyze the contribution of each module of our method, we present ablation experiments in Table~\ref{tab:ablation_CR}. The baseline model only calculates the loss $\mathcal{L}_{pBCE}$ for labeled pixels of the weak view, as described in Section~\ref{Preliminary}. The cross-set data augmentation module brings an average Dice improvement of 4.23\%. Further adding the prototype-based distribution alignment module, our method achieves an average Dice improvement of 5.12\% compared to the baseline. We also conduct ablation experiments on the large-scale LSPL dataset. In Table~\ref{tab:ablation_LSPL}, we present test results on the LSPL dataset and generalization results on two external datasets, BTCV and AbdomenCT-1K. The CDA is demonstrated to be effective in improving the performance of the baseline model with average Dice increments of 2.67\%, 3.11\%, and 2.00\% on the LSPL, BTCV, and AbdomenCT-1K datasets, respectively. Furthermore, the comparison results show that the combination of CDA and PDA promotes the performance of the baseline model, with Dice improvements of 3.01\%, 3.62\%, and 3.04\% on the LSPL, BTCV, and AbdomenCT-1K datasets, respectively. Both of the proposed modules play crucial roles in improving performance.

Fig.~\ref{fig:schematic illustration} shows the distribution of labeled and unlabeled pixels for different categories, visualized using t-SNE~\citep{van2008visualizing}. The baseline model exhibits significant deviation between the labeled and unlabeled pixels for the foreground classes, with considerable confusion between the unlabeled foreground and background. Incorporating CDA consistency regularization reduces the distribution discrepancy between the unlabeled and labeled pixels. Furthermore, with the inclusion of the PDA module, the originally scattered unlabeled foreground data in the background is attracted toward the boundary. Furthermore, we have calculated the intra-class and inter-class variance of the feature embedding. As shown in Table~\ref{tab:Var}, with the help of CDA and PDA, the intra-class variance of the background class and the foreground classes is reduced (except for the kidney category), while the inter-class variance is increased.

\begin{figure}[tbp]
\centering
\setlength{\abovecaptionskip}{-0.3cm}
\includegraphics[width=0.48\textwidth]{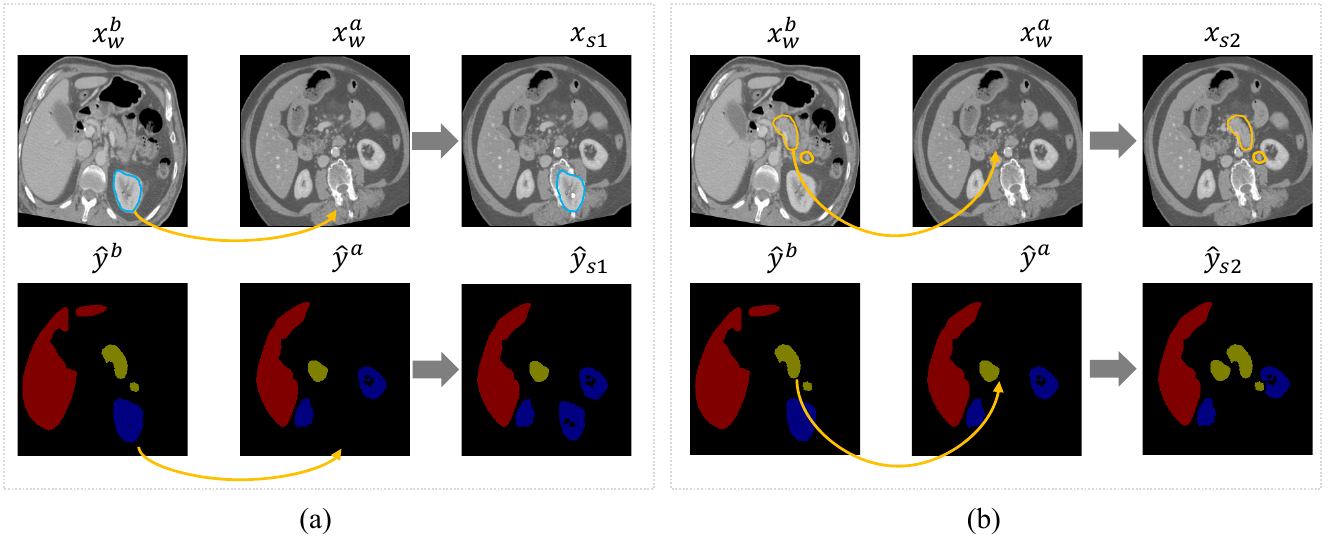}
\caption{Visualizations of differently strongly augmented images generated by CarveMix. (a) and (b) paste the cropped ROI from $x_w^b$ to the same position in $x_w^a$. (a) and (b) crop the ROI of different foreground organs.}
\label{fig:carvemix_strong}
\end{figure}

\begin{figure}[tbp]
    \centering
    \setlength{\abovecaptionskip}{-0.5cm}
    \includegraphics[width=0.48\textwidth]{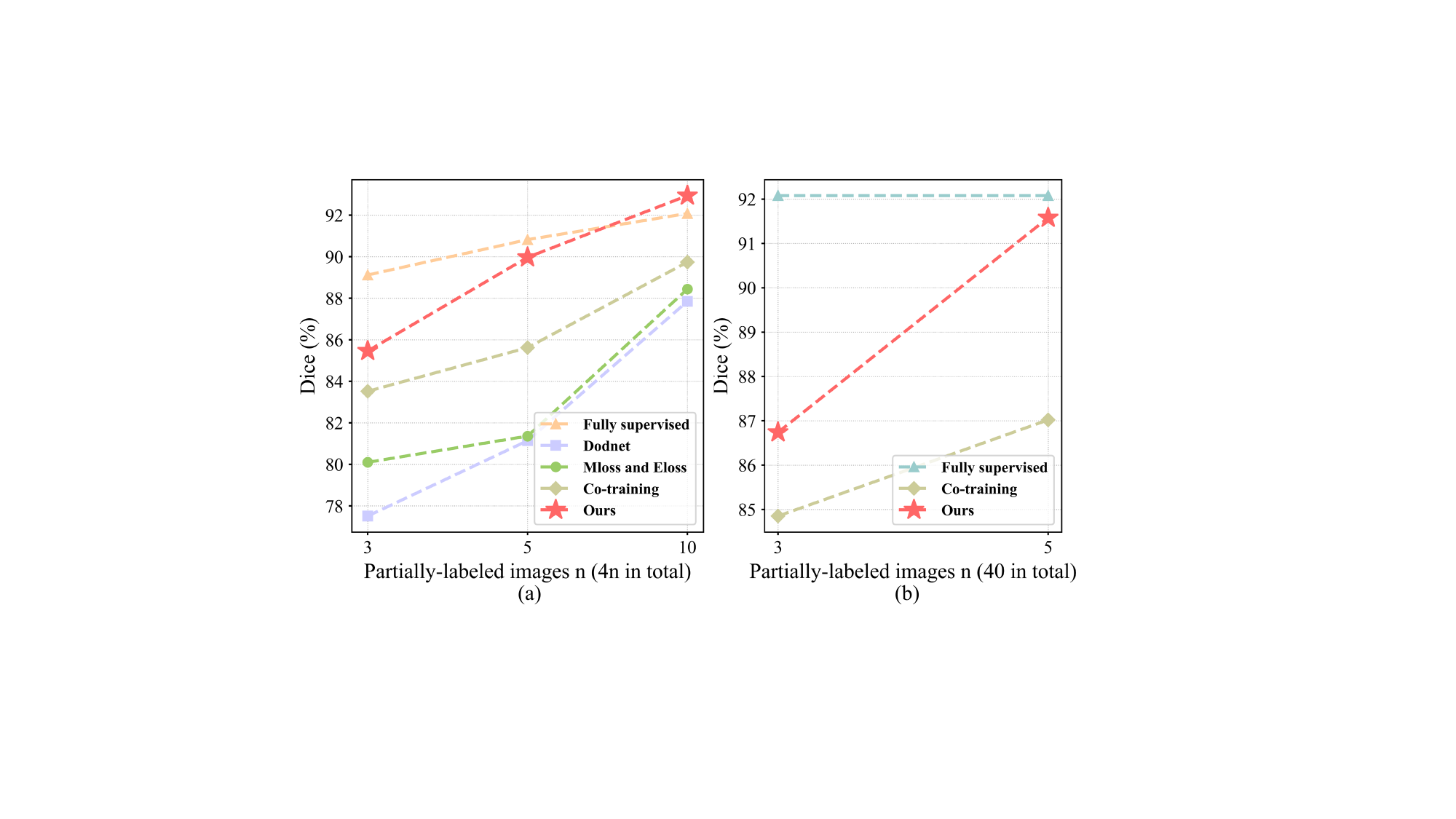}
	\caption{(a) The performance of learning from partially labeled data under different labeled data sizes. (b) The performance of learning from partially labeled data and unlabeled data under different labeled data sizes. }
	\label{ablation_fig2}
\end{figure}

\myparagraph{Comparison of intra-set and cross-set data augmentation.} The design of the augmentation strategy is crucial for consistency regularization. We evaluate the performance of two common strong perturbation strategies in consistency regularization methods, cutout~\citep{devries2017improved} and color jitter~\citep{chen2020simple}. The augmented samples generated by color jitter and cutout are still within the distribution of the original data (intra-set). The cross-set data augmentation strategy performs region mixing between labeled and unlabeled pixels. We compare three cross-set data augmentation strategies, namely CutMix, CowMix, and CarveMix. In Fig.~\ref{fig:carvemix_strong}, we present the augmented images produced using CarveMix. CarveMix involves carving out the ROI of the foreground organs, with the ROI size being sampled from a probability distribution. The ROI is subsequently pasted into the same position of the target image. In Table~\ref{tab:ablation_DA}, we compare the performance of the different data augmentation strategies without using the prototype-based alignment module. As shown in Table~\ref{tab:ablation_DA}, the cross-set strategies achieve comparable performance and consistently outperform the intra-set strategies to a significant extent. This demonstrates that the application of cross-set strong data augmentation can lead to significant improvements over intra-set strong data augmentation. Moreover, our results suggest that the observed improvement stems from the cross-set strategy itself, which effectively reduces the distribution bias, rather than relying solely on a specific augmentation method. 
\begin{table*}[ht]
\scriptsize
\renewcommand\arraystretch{1.2}
\setlength{\belowcaptionskip}{4pt}
\caption{Comparison of different paste position strategies for CutMix.}
\label{tab:cutmix}
\centering
\resizebox{\linewidth}{!}{
\begin{tabular}
{l| ccccc|ccccc}
\shline
\multirow{2}{*}{Method}&
\multicolumn{5}{c|}{Dice(\%) $\uparrow$} & \multicolumn{5}{c}{HD $\downarrow$} 
\\ \cline{2-6} \cline{7-11}  & liver  & spleen  & kidney & pancreas  & avg & liver  & spleen  & kidney & pancreas  & avg  \\ 
\hline
CutMix (same position placement) & 98.29 & 97.71 & 94.57 & 77.63 & 92.05 & 9.66 & 2.87 & 14.47 & 15.18 & 10.55 \\ 
CutMix (random placement) &98.21  &97.52  &95.09  &76.17  &91.74  &10.19  &3.26  &13.00  &14.74  &10.30  \\ 
CutMix (prior position placement) & 98.27 &97.51  &94.99  &77.21  &91.99  &9.56 &3.22  &13.42  &14.51  &10.18 \\ 
\shline
\end{tabular}
}
\end{table*}

\myparagraph{The impact of CutMix’s placement strategy.} We conducted a comparative analysis of three distinct settings: 1) Same position placement. The cropped patch is pasted onto the target image at the same location. 2) Random placement. The cropped patch is randomly placed anywhere on the target image. 3) Prior position placement. The prior position information of each organ is computed by averaging the ground truth of each organ. We impose a constraint that the center position of the pasted cropped patch should be within the prior atlas map. The experimental results in Table~\ref{tab:cutmix} indicate that the performance of these three pasting methods is comparable, with the same position placement and prior position placement methods exhibiting slightly better results than random placement. 

\begin{figure}[t]
    \centering
    \setlength{\abovecaptionskip}{0.0cm}
    \textbf{\includegraphics[width=0.3\textwidth]{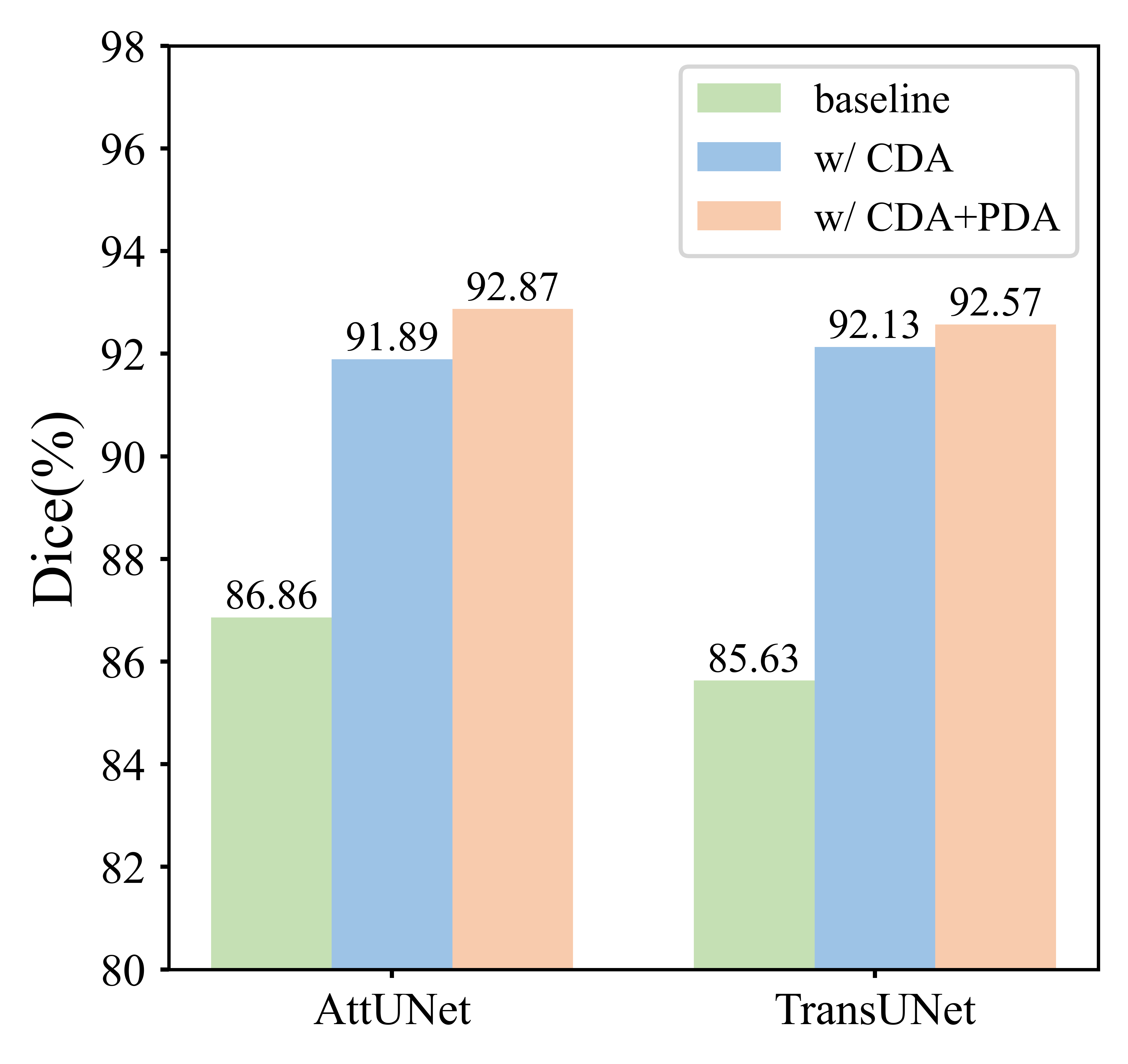}}
	\caption{Ablation results of different segmentation backbones.}
	\label{ablation_fig3}
\end{figure}

\myparagraph{The impact of the number of prototypes.} In prototype-based classifiers, each class can be represented by a set of prototypes, providing comprehensive class representations and capturing intra-class variations. To investigate the impact of the number of prototypes on model performance, we conduct experiments on the toy dataset and report the average Dice of the four organs in Fig.~\ref{ablation_fig1}(a). With one prototype per class ($K=1$), the average Dice is 92.61\%. As we increase the number of prototypes ($K=5$), the performance improves to 92.94\%. However, when the number of prototypes exceeds 5, there is no further performance gain. Consequently, we set the number of prototypes to 5.

\myparagraph{The impact of the number of strong views.} We examine the impact of the number $M$ of strongly augmented views on model performance in Fig.~\ref{ablation_fig1}(b). As the number of strong views increases to 2, the performance improves to 92.94\%. However, exceeding this number may lead to a slight decrease in performance, indicating that utilizing two strong views may be adequate to effectively utilize the image-level perturbation space.

\myparagraph{Performance under small-scale data sizes.} We further investigate the impact of the partially labeled dataset size on model performance. In Fig.~\ref{ablation_fig2}(a), we compare the average Dice of different methods under various data sizes, with the performance of the fully-supervised setting serving as a reference. In detail, the size $n$ of each partially-labeled dataset $D_c$ is 3, 5, and 10 respectively (the number of training data ${\{D_c\}}_{c=1}^{4}$ is $4n$, consequently becoming 12, 20, and 40). The experiments in Table~\ref{tab:SOTA_toy} are conducted with $n=10$. Our method consistently delivers robust performance across different data sizes. In particular, in the case of small training data, our method significantly outperforms other methods that are prone to overfitting with limited training data. As the amount of data increases, the performance gap between different methods gradually diminishes, and our method remains the best, even surpassing the performance of the fully-supervised method. As shown in Table~\ref{tab:SOTA_toy}, when $n$ is 10, our method improves the average Dice by 0.86\%  compared to the fully supervised method, with a notable improvement of 2.93\% in the pancreas category. Furthermore, our method and the other pseudo-labeling methods (co-training, self-training, Ms-kd, CPS, and DMPLS) can be easily extended to a semi-supervised learning setting, \ie, learning from partially labeled images and fully unlabeled images. In Fig.~\ref{ablation_fig2}(b), we compare the proposed method with co-training. When the size $n$ of $D_c$ is 3, 5 respectively, the size of fully unlabeled images $ D_u$ is 28, 20, and the size of the training set $D=\{{\{D_c\}}_{c=1}^4, D_u\}$ is always 40. The fully-supervised performance learned from 40 training images is given in Fig.~\ref{ablation_fig2}(b) as a reference. Comparing the results in Fig.~\ref{ablation_fig2}(a) obtained from learning only from partially labeled data, we observe that incorporating fully unlabeled images into the training can bring performance gains, and our method outperforms the co-training method.

\begin{figure}[t]
    \centering
    \setlength{\abovecaptionskip}{0.0cm} \textbf{\includegraphics[width=0.48\textwidth]{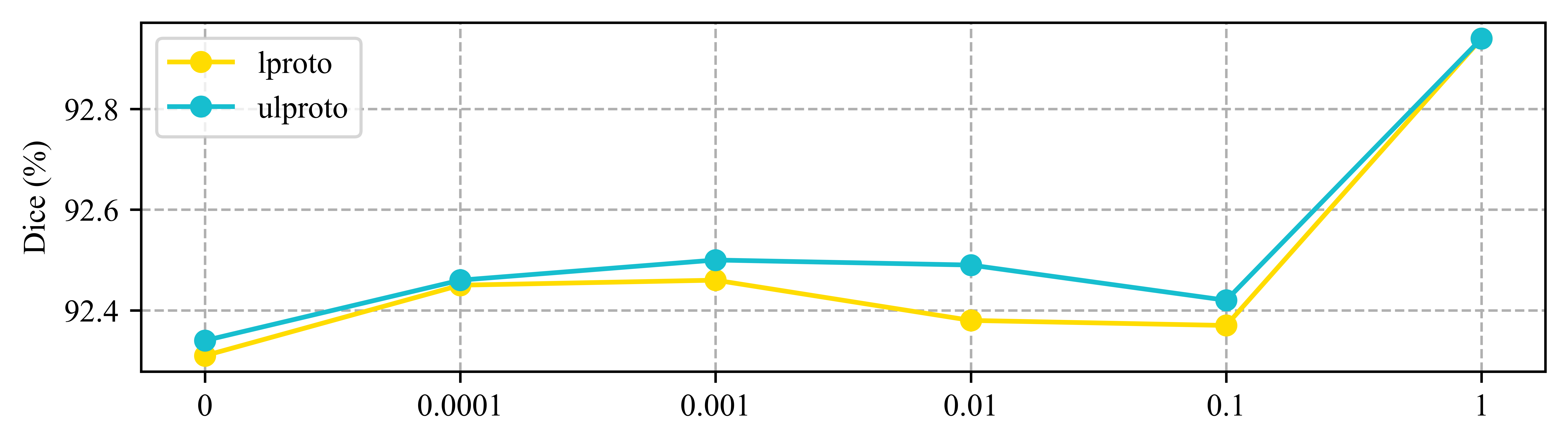}}
    \vspace{-2mm}
	\caption{Ablation study of weight parameters of $\mathcal{L}_{lproto}$ and $\mathcal{L}_{ulproto}$.}
	\label{fig:lprotoulproto}
\end{figure}

\myparagraph{Ablation study of different backbones.} We further apply the proposed method to other segmentation backbones (namely Attention U-Net (AttUNet)~\citep{oktay2018attention} and TransUNet~\citep{chen2021transunet}) to verify the effectiveness of the proposed method. From Fig.~\ref{ablation_fig3}, we can observe that our proposed method can significantly outperform the baseline on various backbones, highlighting its effectiveness and broad applicability.

\begin{figure}[btp]
    \centering
    \setlength{\abovecaptionskip}{0.0cm}
    \textbf{\includegraphics[width=0.48\textwidth]{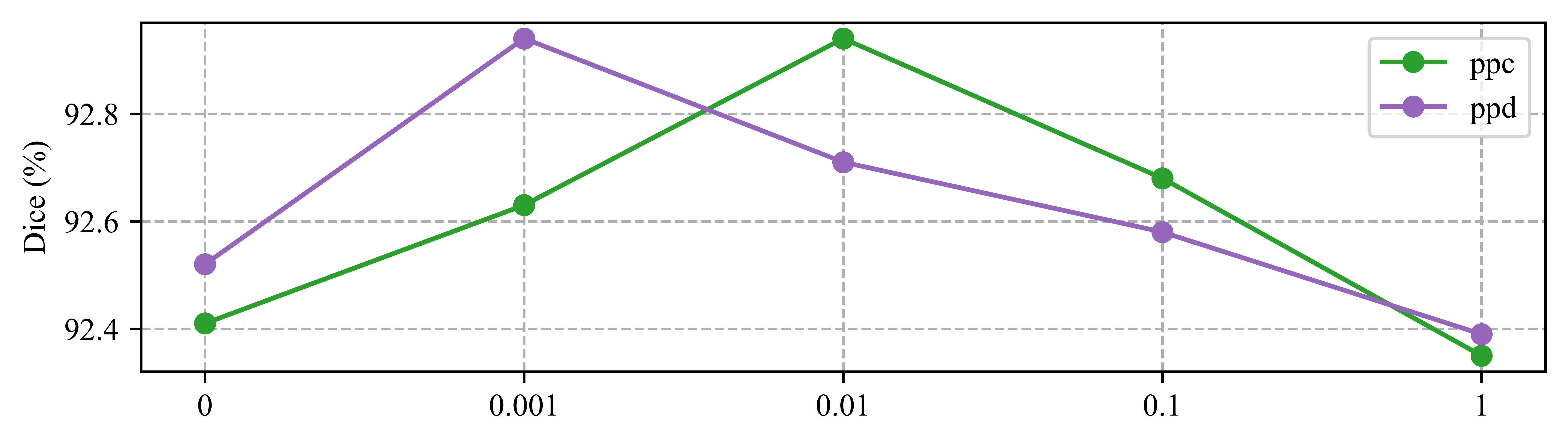}}
    \vspace{-2mm}
	\caption{Ablation study of weight parameters of $\mathcal{L}_{ppd}$ and $\mathcal{L}_{ppc}$.}
	\label{fig:ppcppd}
\end{figure}
\myparagraph{Ablation study of weight parameters.} We investigate the impact of weight parameters of $\mathcal{L}_{lproto}$ and $\mathcal{L}_{ulproto}$ in Equation~\ref{eq:loss_strongview} on the performance. Ablation experiments are conducted whereby one of the parameters is held constant (set to 1) while the other is varied. The results, presented in Fig.~\ref{fig:lprotoulproto}, demonstrate that an optimal performance is achieved when the weights of both $\mathcal{L}_{lproto}$ and $\mathcal{L}_{ulproto}$ are set to 1. In Fig.~\ref{fig:ppcppd}, We performed ablation experiments on the weights $\lambda_1$ and $\lambda_2$ of $\mathcal{L}_{ppd}$ and $\mathcal{L}_{ppc}$. Our results indicate that the optimal combination of parameters is $\lambda_1$=0.001 and $\lambda_2$=0.01.

\begin{table}[tbp]
\large
\renewcommand\arraystretch{1.2}
\setlength{\belowcaptionskip}{4pt}
\caption{Intra-class and inter-class variance for different methods.}
\label{tab:Var}
\centering
\resizebox{\linewidth}{!}{
\begin{tabular}{l|cccccc|c|c}
\shline
\multirow{2}{*}{Method} & \multicolumn{6}{c|}{intra-class variance $\downarrow$} & inter-class & \multirow{2}{*}{$\frac{inter-class}{intra-class}$ $\uparrow$} \\ \cline{2-7}
 & liver & spleen & kidney & pancreas & background & average & variance $\uparrow$ &  \\
 \hline
baseline & 0.0401 & 0.0338 & \textbf{0.0517} & 0.2253 & 0.5043 & 0.1710 & 0.5362 & 3.1352 \\
w/ CDA & 0.0451 & \textbf{0.0162} & 0.0662 & \textbf{0.1227} & 0.5191 & 0.1539 & 0.5593 & 3.6351 \\
Ours & \textbf{0.0377} & 0.0263 & 0.0645 & 0.1395 & \textbf{0.4736} & \textbf{0.1483} & \textbf{0.5663} & \textbf{3.8176} \\
\shline
\end{tabular}
}
\end{table}

\section{Conclusion}
Partially-supervised segmentation is confronted with the challenge of feature distribution mismatch between the labeled and unlabeled pixels. To address this issue, this paper proposes a new framework called LTUDA for partially supervised multi-organ segmentation. Our approach comprises two key components. Firstly, we design a cross-set data augmentation strategy that generates interpolated samples between the labeled and unlabeled pixels, thereby reducing feature discrepancy. Furthermore, we propose a prototype-based distribution alignment module to align the distributions of labeled and unlabeled pixels. This module eliminates confusion between unlabeled foreground and background by encouraging consistency between labeled prototype and unlabeled prototype classifier. Experimental results on both a toy dataset and a large-scale partially labeled dataset demonstrate the effectiveness of our method.

Our proposed method has the potential to contribute to the development of foundation models. Recent foundation models~\citep{liu2023clip,ye2023uniseg} have focused on leveraging large-scale and diverse partially annotated datasets to segment different types of organs and tumors, thereby fully exploiting existing publicly available datasets from different modalities (\eg, CT, MRI, PET). While these models have achieved promising results, they typically do not leverage unlabeled pixels or consider the domain shift between partially-labeled datasets from different institutions. This heterogeneity in training data leads to a more pronounced distributional deviation between labeled and unlabeled pixels, posing a greater challenge. Our proposed labeled-to-unlabeled distribution framework specifically addresses this challenge and thus can help unlock the potential of public datasets. Looking ahead, one promising future direction is to apply partially supervised segmentation methods to more realistic clinical scenarios ~\citep{yao2021deepprognosis}. Our approach demonstrates the feasibility of training multi-organ segmentation models using small-sized partially labeled datasets. This convenience enables the customization and development of multi-organ segmentation models based on the clinical requirements of medical professionals. If the annotations provided by the public datasets lack specific anatomical structures of interest, experts would only need to provide annotations of a small number of missing classes to develop multi-organ segmentation models.

\section*{Acknowledgments}
This work was supported by the National Natural Science Foundation of China / Research Grants Council Joint Research Scheme
under Grant N\_HKUST627/20, the National Natural Science Foundation of China (62061160490 and U20B2064).

\bibliographystyle{model2-names.bst}
\biboptions{authoryear}
\bibliography{main}

\begin{thebibliography}{82}
\expandafter\ifx\csname natexlab\endcsname\relax\def\natexlab#1{#1}\fi
\providecommand{\url}[1]{\texttt{#1}}
\providecommand{\href}[2]{#2}
\providecommand{\path}[1]{#1}
\providecommand{\DOIprefix}{doi:}
\providecommand{\ArXivprefix}{arXiv:}
\providecommand{\URLprefix}{URL: }
\providecommand{\Pubmedprefix}{pmid:}
\providecommand{\doi}[1]{\href{http://dx.doi.org/#1}{\path{#1}}}
\providecommand{\Pubmed}[1]{\href{pmid:#1}{\path{#1}}}
\providecommand{\bibinfo}[2]{#2}
\ifx\xfnm\relax \def\xfnm[#1]{\unskip,\space#1}\fi
\bibitem[{Abdelfattah et~al.(2022)Abdelfattah, Zhang, Wu, Wu, Wang and Wang}]{abdelfattah2022plmcl}
\bibinfo{author}{Abdelfattah, R.}, \bibinfo{author}{Zhang, X.}, \bibinfo{author}{Wu, Z.}, \bibinfo{author}{Wu, X.}, \bibinfo{author}{Wang, X.}, \bibinfo{author}{Wang, S.}, \bibinfo{year}{2022}.
\newblock \bibinfo{title}{Plmcl: Partial-label momentum curriculum learning for multi-label image classification}.
\newblock \bibinfo{journal}{arXiv preprint arXiv:2208.09999} .
\bibitem[{Berthelot et~al.(2019)Berthelot, Carlini, Cubuk, Kurakin, Sohn, Zhang and Raffel}]{berthelot2019remixmatch}
\bibinfo{author}{Berthelot, D.}, \bibinfo{author}{Carlini, N.}, \bibinfo{author}{Cubuk, E.D.}, \bibinfo{author}{Kurakin, A.}, \bibinfo{author}{Sohn, K.}, \bibinfo{author}{Zhang, H.}, \bibinfo{author}{Raffel, C.}, \bibinfo{year}{2019}.
\newblock \bibinfo{title}{Remixmatch: Semi-supervised learning with distribution alignment and augmentation anchoring}.
\newblock \bibinfo{journal}{arXiv preprint arXiv:1911.09785} .
\bibitem[{Bilic et~al.(2023)Bilic, Christ, Li, Vorontsov, Ben-Cohen, Kaissis, Szeskin, Jacobs, Mamani, Chartrand et~al.}]{bilic2023liver}
\bibinfo{author}{Bilic, P.}, \bibinfo{author}{Christ, P.}, \bibinfo{author}{Li, H.B.}, \bibinfo{author}{Vorontsov, E.}, \bibinfo{author}{Ben-Cohen, A.}, \bibinfo{author}{Kaissis, G.}, \bibinfo{author}{Szeskin, A.}, \bibinfo{author}{Jacobs, C.}, \bibinfo{author}{Mamani, G.E.H.}, \bibinfo{author}{Chartrand, G.}, et~al., \bibinfo{year}{2023}.
\newblock \bibinfo{title}{The liver tumor segmentation benchmark (lits)}.
\newblock \bibinfo{journal}{Medical Image Analysis} \bibinfo{volume}{84}, \bibinfo{pages}{102680}.
\bibitem[{Cao et~al.(2020)Cao, Chen, Li, Peng, Wang and Cheng}]{cao2020uncertainty}
\bibinfo{author}{Cao, X.}, \bibinfo{author}{Chen, H.}, \bibinfo{author}{Li, Y.}, \bibinfo{author}{Peng, Y.}, \bibinfo{author}{Wang, S.}, \bibinfo{author}{Cheng, L.}, \bibinfo{year}{2020}.
\newblock \bibinfo{title}{Uncertainty aware temporal-ensembling model for semi-supervised abus mass segmentation}.
\newblock \bibinfo{journal}{IEEE transactions on medical imaging} \bibinfo{volume}{40}, \bibinfo{pages}{431--443}.
\bibitem[{Caron et~al.(2020)Caron, Misra, Mairal, Goyal, Bojanowski and Joulin}]{caron2020unsupervised}
\bibinfo{author}{Caron, M.}, \bibinfo{author}{Misra, I.}, \bibinfo{author}{Mairal, J.}, \bibinfo{author}{Goyal, P.}, \bibinfo{author}{Bojanowski, P.}, \bibinfo{author}{Joulin, A.}, \bibinfo{year}{2020}.
\newblock \bibinfo{title}{Unsupervised learning of visual features by contrasting cluster assignments}.
\newblock \bibinfo{journal}{Advances in Neural Information Processing Systems} \bibinfo{volume}{33}, \bibinfo{pages}{9912--9924}.
\bibitem[{Cerrolaza et~al.(2019)Cerrolaza, Picazo, Humbert, Sato, Rueckert, Ballester and Linguraru}]{cerrolaza2019computational}
\bibinfo{author}{Cerrolaza, J.J.}, \bibinfo{author}{Picazo, M.L.}, \bibinfo{author}{Humbert, L.}, \bibinfo{author}{Sato, Y.}, \bibinfo{author}{Rueckert, D.}, \bibinfo{author}{Ballester, M.{\'A}.G.}, \bibinfo{author}{Linguraru, M.G.}, \bibinfo{year}{2019}.
\newblock \bibinfo{title}{Computational anatomy for multi-organ analysis in medical imaging: A review}.
\newblock \bibinfo{journal}{Medical Image Analysis} \bibinfo{volume}{56}, \bibinfo{pages}{44--67}.
\bibitem[{Chen et~al.(2022a)Chen, Jiang, Wang, Wan, Wang and Long}]{chen2022debiased}
\bibinfo{author}{Chen, B.}, \bibinfo{author}{Jiang, J.}, \bibinfo{author}{Wang, X.}, \bibinfo{author}{Wan, P.}, \bibinfo{author}{Wang, J.}, \bibinfo{author}{Long, M.}, \bibinfo{year}{2022}a.
\newblock \bibinfo{title}{Debiased self-training for semi-supervised learning}.
\newblock \bibinfo{journal}{Advances in Neural Information Processing Systems} \bibinfo{volume}{35}, \bibinfo{pages}{32424--32437}.
\bibitem[{Chen et~al.(2021a)Chen, Lu, Yu, Luo, Adeli, Wang, Lu, Yuille and Zhou}]{chen2021transunet}
\bibinfo{author}{Chen, J.}, \bibinfo{author}{Lu, Y.}, \bibinfo{author}{Yu, Q.}, \bibinfo{author}{Luo, X.}, \bibinfo{author}{Adeli, E.}, \bibinfo{author}{Wang, Y.}, \bibinfo{author}{Lu, L.}, \bibinfo{author}{Yuille, A.L.}, \bibinfo{author}{Zhou, Y.}, \bibinfo{year}{2021}a.
\newblock \bibinfo{title}{Transunet: Transformers make strong encoders for medical image segmentation}.
\newblock \bibinfo{journal}{arXiv preprint arXiv:2102.04306} .
\bibitem[{Chen et~al.(2019)Chen, Ma and Zheng}]{chen2019med3d}
\bibinfo{author}{Chen, S.}, \bibinfo{author}{Ma, K.}, \bibinfo{author}{Zheng, Y.}, \bibinfo{year}{2019}.
\newblock \bibinfo{title}{Med3d: Transfer learning for 3d medical image analysis}.
\newblock \bibinfo{journal}{arXiv preprint arXiv:1904.00625} .
\bibitem[{Chen et~al.(2020)Chen, Kornblith, Norouzi and Hinton}]{chen2020simple}
\bibinfo{author}{Chen, T.}, \bibinfo{author}{Kornblith, S.}, \bibinfo{author}{Norouzi, M.}, \bibinfo{author}{Hinton, G.}, \bibinfo{year}{2020}.
\newblock \bibinfo{title}{A simple framework for contrastive learning of visual representations}, in: \bibinfo{booktitle}{International Conference on Machine Learning}, pp. \bibinfo{pages}{1597--1607}.
\bibitem[{Chen et~al.(2022b)Chen, Pu, Wu, Xie and Lin}]{chen2022structured}
\bibinfo{author}{Chen, T.}, \bibinfo{author}{Pu, T.}, \bibinfo{author}{Wu, H.}, \bibinfo{author}{Xie, Y.}, \bibinfo{author}{Lin, L.}, \bibinfo{year}{2022}b.
\newblock \bibinfo{title}{Structured semantic transfer for multi-label recognition with partial labels}, in: \bibinfo{booktitle}{Proceedings of the AAAI Conference on Artificial Intelligence}, pp. \bibinfo{pages}{339--346}.
\bibitem[{Chen et~al.(2021b)Chen, Yuan, Zeng and Wang}]{chen2021semi}
\bibinfo{author}{Chen, X.}, \bibinfo{author}{Yuan, Y.}, \bibinfo{author}{Zeng, G.}, \bibinfo{author}{Wang, J.}, \bibinfo{year}{2021}b.
\newblock \bibinfo{title}{Semi-supervised semantic segmentation with cross pseudo supervision}, in: \bibinfo{booktitle}{Proceedings of the IEEE/CVF Conference on Computer Vision and Pattern Recognition}, pp. \bibinfo{pages}{2613--2622}.
\bibitem[{Chen et~al.(2023)Chen, Yang and Bai}]{chen2023confidence}
\bibinfo{author}{Chen, Y.}, \bibinfo{author}{Yang, X.}, \bibinfo{author}{Bai, X.}, \bibinfo{year}{2023}.
\newblock \bibinfo{title}{Confidence-weighted mutual supervision on dual networks for unsupervised cross-modality image segmentation}.
\newblock \bibinfo{journal}{Science China Information Sciences} \bibinfo{volume}{66}, \bibinfo{pages}{210104}.
\bibitem[{Chu et~al.(2013)Chu, Oda, Kitasaka, Misawa, Fujiwara, Hayashi, Nimura, Rueckert and Mori}]{chu2013multi}
\bibinfo{author}{Chu, C.}, \bibinfo{author}{Oda, M.}, \bibinfo{author}{Kitasaka, T.}, \bibinfo{author}{Misawa, K.}, \bibinfo{author}{Fujiwara, M.}, \bibinfo{author}{Hayashi, Y.}, \bibinfo{author}{Nimura, Y.}, \bibinfo{author}{Rueckert, D.}, \bibinfo{author}{Mori, K.}, \bibinfo{year}{2013}.
\newblock \bibinfo{title}{Multi-organ segmentation based on spatially-divided probabilistic atlas from 3d abdominal ct images}, in: \bibinfo{booktitle}{International Conference on Medical Image Computing and Computer-Assisted Intervention}, pp. \bibinfo{pages}{165--172}.
\bibitem[{DeVries and Taylor(2017)}]{devries2017improved}
\bibinfo{author}{DeVries, T.}, \bibinfo{author}{Taylor, G.W.}, \bibinfo{year}{2017}.
\newblock \bibinfo{title}{Improved regularization of convolutional neural networks with cutout}.
\newblock \bibinfo{journal}{arXiv preprint arXiv:1708.04552} .
\bibitem[{Dmitriev and Kaufman(2019)}]{dmitriev2019learning}
\bibinfo{author}{Dmitriev, K.}, \bibinfo{author}{Kaufman, A.E.}, \bibinfo{year}{2019}.
\newblock \bibinfo{title}{Learning multi-class segmentations from single-class datasets}, in: \bibinfo{booktitle}{Proceedings of the IEEE/CVF Conference on Computer Vision and Pattern Recognition}, pp. \bibinfo{pages}{9501--9511}.
\bibitem[{Dong et~al.(2022)Dong, Kampffmeyer, Liang, Xu, Voiculescu and Xing}]{dong2022towards}
\bibinfo{author}{Dong, N.}, \bibinfo{author}{Kampffmeyer, M.}, \bibinfo{author}{Liang, X.}, \bibinfo{author}{Xu, M.}, \bibinfo{author}{Voiculescu, I.}, \bibinfo{author}{Xing, E.}, \bibinfo{year}{2022}.
\newblock \bibinfo{title}{Towards robust partially supervised multi-structure medical image segmentation on small-scale data}.
\newblock \bibinfo{journal}{Applied Soft Computing} \bibinfo{volume}{114}, \bibinfo{pages}{108074}.
\bibitem[{Dosovitskiy et~al.(2020)Dosovitskiy, Beyer, Kolesnikov, Weissenborn, Zhai, Unterthiner, Dehghani, Minderer, Heigold, Gelly et~al.}]{dosovitskiy2020image}
\bibinfo{author}{Dosovitskiy, A.}, \bibinfo{author}{Beyer, L.}, \bibinfo{author}{Kolesnikov, A.}, \bibinfo{author}{Weissenborn, D.}, \bibinfo{author}{Zhai, X.}, \bibinfo{author}{Unterthiner, T.}, \bibinfo{author}{Dehghani, M.}, \bibinfo{author}{Minderer, M.}, \bibinfo{author}{Heigold, G.}, \bibinfo{author}{Gelly, S.}, et~al., \bibinfo{year}{2020}.
\newblock \bibinfo{title}{An image is worth 16x16 words: Transformers for image recognition at scale}.
\newblock \bibinfo{journal}{arXiv preprint arXiv:2010.11929} .
\bibitem[{Durand et~al.(2019)Durand, Mehrasa and Mori}]{durand2019learning}
\bibinfo{author}{Durand, T.}, \bibinfo{author}{Mehrasa, N.}, \bibinfo{author}{Mori, G.}, \bibinfo{year}{2019}.
\newblock \bibinfo{title}{Learning a deep convnet for multi-label classification with partial labels}, in: \bibinfo{booktitle}{Proceedings of the IEEE/CVF Conference on Computer Vision and Pattern Recognition}, pp. \bibinfo{pages}{647--657}.
\bibitem[{Fang and Yan(2020)}]{fang2020multi}
\bibinfo{author}{Fang, X.}, \bibinfo{author}{Yan, P.}, \bibinfo{year}{2020}.
\newblock \bibinfo{title}{Multi-organ segmentation over partially labeled datasets with multi-scale feature abstraction}.
\newblock \bibinfo{journal}{IEEE Transactions on Medical Imaging} \bibinfo{volume}{39}, \bibinfo{pages}{3619--3629}.
\bibitem[{Feng et~al.(2021)Feng, Zhou, Zhang, Zhang and Wang}]{feng2021ms}
\bibinfo{author}{Feng, S.}, \bibinfo{author}{Zhou, Y.}, \bibinfo{author}{Zhang, X.}, \bibinfo{author}{Zhang, Y.}, \bibinfo{author}{Wang, Y.}, \bibinfo{year}{2021}.
\newblock \bibinfo{title}{Ms-kd: Multi-organ segmentation with multiple binary-labeled datasets}.
\newblock \bibinfo{journal}{arXiv preprint arXiv:2108.02559} .
\bibitem[{French et~al.(2019)French, Aila, Laine, Mackiewicz and Finlayson}]{french2019semi}
\bibinfo{author}{French, G.}, \bibinfo{author}{Aila, T.}, \bibinfo{author}{Laine, S.}, \bibinfo{author}{Mackiewicz, M.}, \bibinfo{author}{Finlayson, G.}, \bibinfo{year}{2019}.
\newblock \bibinfo{title}{Semi-supervised semantic segmentation needs strong, high-dimensional perturbations}.
\newblock \bibinfo{journal}{British Machine Vision Conference(BMVC)} .
\bibitem[{French et~al.(2020)French, Oliver and Salimans}]{french2020milking}
\bibinfo{author}{French, G.}, \bibinfo{author}{Oliver, A.}, \bibinfo{author}{Salimans, T.}, \bibinfo{year}{2020}.
\newblock \bibinfo{title}{Milking cowmask for semi-supervised image classification}.
\newblock \bibinfo{journal}{arXiv preprint arXiv:2003.12022} .
\bibitem[{Gibson et~al.(2018)Gibson, Giganti, Hu, Bonmati, Bandula, Gurusamy, Davidson, Pereira, Clarkson and Barratt}]{gibson2018automatic}
\bibinfo{author}{Gibson, E.}, \bibinfo{author}{Giganti, F.}, \bibinfo{author}{Hu, Y.}, \bibinfo{author}{Bonmati, E.}, \bibinfo{author}{Bandula, S.}, \bibinfo{author}{Gurusamy, K.}, \bibinfo{author}{Davidson, B.}, \bibinfo{author}{Pereira, S.P.}, \bibinfo{author}{Clarkson, M.J.}, \bibinfo{author}{Barratt, D.C.}, \bibinfo{year}{2018}.
\newblock \bibinfo{title}{Automatic multi-organ segmentation on abdominal ct with dense v-networks}.
\newblock \bibinfo{journal}{IEEE Transactions on Medical Imaging} \bibinfo{volume}{37}, \bibinfo{pages}{1822--1834}.
\bibitem[{G{\'o}mez et~al.(2020)G{\'o}mez, Mesejo, Ib{\'a}{\~n}ez, Valsecchi and Cord{\'o}n}]{gomez2020deep}
\bibinfo{author}{G{\'o}mez, O.}, \bibinfo{author}{Mesejo, P.}, \bibinfo{author}{Ib{\'a}{\~n}ez, O.}, \bibinfo{author}{Valsecchi, A.}, \bibinfo{author}{Cord{\'o}n, O.}, \bibinfo{year}{2020}.
\newblock \bibinfo{title}{Deep architectures for high-resolution multi-organ chest x-ray image segmentation}.
\newblock \bibinfo{journal}{Neural Computing and Applications} \bibinfo{volume}{32}, \bibinfo{pages}{15949--15963}.
\bibitem[{Guo et~al.(2021)Guo, Liu and Yuan}]{guo2021semantic}
\bibinfo{author}{Guo, X.}, \bibinfo{author}{Liu, J.}, \bibinfo{author}{Yuan, Y.}, \bibinfo{year}{2021}.
\newblock \bibinfo{title}{Semantic-oriented labeled-to-unlabeled distribution translation for image segmentation}.
\newblock \bibinfo{journal}{IEEE transactions on medical imaging} \bibinfo{volume}{41}, \bibinfo{pages}{434--445}.
\bibitem[{He et~al.(2016)He, Zhang, Ren and Sun}]{he2016deep}
\bibinfo{author}{He, K.}, \bibinfo{author}{Zhang, X.}, \bibinfo{author}{Ren, S.}, \bibinfo{author}{Sun, J.}, \bibinfo{year}{2016}.
\newblock \bibinfo{title}{Deep residual learning for image recognition}, in: \bibinfo{booktitle}{Proceedings of the IEEE/CVF Conference on Computer Vision and Pattern Recognition}, pp. \bibinfo{pages}{770--778}.
\bibitem[{Heller et~al.(2019)Heller, Sathianathen, Kalapara, Walczak, Moore, Kaluzniak, Rosenberg, Blake, Rengel, Oestreich et~al.}]{heller2019kits19}
\bibinfo{author}{Heller, N.}, \bibinfo{author}{Sathianathen, N.}, \bibinfo{author}{Kalapara, A.}, \bibinfo{author}{Walczak, E.}, \bibinfo{author}{Moore, K.}, \bibinfo{author}{Kaluzniak, H.}, \bibinfo{author}{Rosenberg, J.}, \bibinfo{author}{Blake, P.}, \bibinfo{author}{Rengel, Z.}, \bibinfo{author}{Oestreich, M.}, et~al., \bibinfo{year}{2019}.
\newblock \bibinfo{title}{The kits19 challenge data: 300 kidney tumor cases with clinical context, ct semantic segmentations, and surgical outcomes}.
\newblock \bibinfo{journal}{arXiv preprint arXiv:1904.00445} .
\bibitem[{Hendrycks and Gimpel(2016)}]{hendrycks2016baseline}
\bibinfo{author}{Hendrycks, D.}, \bibinfo{author}{Gimpel, K.}, \bibinfo{year}{2016}.
\newblock \bibinfo{title}{A baseline for detecting misclassified and out-of-distribution examples in neural networks}.
\newblock \bibinfo{journal}{arXiv preprint arXiv:1610.02136} .
\bibitem[{Huang et~al.(2020)Huang, Zheng, Hu, Zhang and Li}]{huang2020multi}
\bibinfo{author}{Huang, R.}, \bibinfo{author}{Zheng, Y.}, \bibinfo{author}{Hu, Z.}, \bibinfo{author}{Zhang, S.}, \bibinfo{author}{Li, H.}, \bibinfo{year}{2020}.
\newblock \bibinfo{title}{Multi-organ segmentation via co-training weight-averaged models from few-organ datasets}, in: \bibinfo{booktitle}{International Conference on Medical Image Computing and Computer-Assisted Intervention}, pp. \bibinfo{pages}{146--155}.
\bibitem[{Isensee et~al.(2021)Isensee, Jaeger, Kohl, Petersen and Maier-Hein}]{isensee2021nnu}
\bibinfo{author}{Isensee, F.}, \bibinfo{author}{Jaeger, P.F.}, \bibinfo{author}{Kohl, S.A.}, \bibinfo{author}{Petersen, J.}, \bibinfo{author}{Maier-Hein, K.H.}, \bibinfo{year}{2021}.
\newblock \bibinfo{title}{nnu-net: a self-configuring method for deep learning-based biomedical image segmentation}.
\newblock \bibinfo{journal}{Nature methods} \bibinfo{volume}{18}, \bibinfo{pages}{203--211}.
\bibitem[{Kang et~al.(2021)Kang, Lu, Yuille and Zhou}]{kang2021label}
\bibinfo{author}{Kang, M.}, \bibinfo{author}{Lu, Y.}, \bibinfo{author}{Yuille, A.L.}, \bibinfo{author}{Zhou, Z.}, \bibinfo{year}{2021}.
\newblock \bibinfo{title}{Label-assemble: Leveraging multiple datasets with partial labels}.
\newblock \bibinfo{journal}{arXiv preprint arXiv:2109.12265} .
\bibitem[{Ke et~al.(2020)Ke, Qiu, Li, Yan and Lau}]{ke2020guided}
\bibinfo{author}{Ke, Z.}, \bibinfo{author}{Qiu, D.}, \bibinfo{author}{Li, K.}, \bibinfo{author}{Yan, Q.}, \bibinfo{author}{Lau, R.W.}, \bibinfo{year}{2020}.
\newblock \bibinfo{title}{Guided collaborative training for pixel-wise semi-supervised learning}, in: \bibinfo{booktitle}{Computer Vision--ECCV 2020: 16th European Conference, Glasgow, UK, August 23--28, 2020, Proceedings, Part XIII 16}, \bibinfo{organization}{Springer}. pp. \bibinfo{pages}{429--445}.
\bibitem[{Kim et~al.(2020)Kim, Jang, Park and Jeong}]{kim2020structured}
\bibinfo{author}{Kim, J.}, \bibinfo{author}{Jang, J.}, \bibinfo{author}{Park, H.}, \bibinfo{author}{Jeong, S.}, \bibinfo{year}{2020}.
\newblock \bibinfo{title}{Structured consistency loss for semi-supervised semantic segmentation}.
\newblock \bibinfo{journal}{arXiv preprint arXiv:2001.04647} .
\bibitem[{Kim et~al.(2022)Kim, Kim, Akata and Lee}]{kim2022large}
\bibinfo{author}{Kim, Y.}, \bibinfo{author}{Kim, J.M.}, \bibinfo{author}{Akata, Z.}, \bibinfo{author}{Lee, J.}, \bibinfo{year}{2022}.
\newblock \bibinfo{title}{Large loss matters in weakly supervised multi-label classification}, in: \bibinfo{booktitle}{Proceedings of the IEEE/CVF Conference on Computer Vision and Pattern Recognition}, pp. \bibinfo{pages}{14156--14165}.
\bibitem[{Kumar et~al.(2019)Kumar, Verma, Anand, Zhou, Onder, Tsougenis, Chen, Heng, Li, Hu et~al.}]{kumar2019multi}
\bibinfo{author}{Kumar, N.}, \bibinfo{author}{Verma, R.}, \bibinfo{author}{Anand, D.}, \bibinfo{author}{Zhou, Y.}, \bibinfo{author}{Onder, O.F.}, \bibinfo{author}{Tsougenis, E.}, \bibinfo{author}{Chen, H.}, \bibinfo{author}{Heng, P.A.}, \bibinfo{author}{Li, J.}, \bibinfo{author}{Hu, Z.}, et~al., \bibinfo{year}{2019}.
\newblock \bibinfo{title}{A multi-organ nucleus segmentation challenge}.
\newblock \bibinfo{journal}{IEEE Transactions on Medical Imaging} \bibinfo{volume}{39}, \bibinfo{pages}{1380--1391}.
\bibitem[{Landman et~al.(2017)Landman, Xu, Igelsias, Styner, Langerak and Klein}]{landman2017multi}
\bibinfo{author}{Landman, B.}, \bibinfo{author}{Xu, Z.}, \bibinfo{author}{Igelsias, J.E.}, \bibinfo{author}{Styner, M.}, \bibinfo{author}{Langerak, T.R.}, \bibinfo{author}{Klein, A.}, \bibinfo{year}{2017}.
\newblock \bibinfo{title}{Multi-atlas labeling beyond the cranial vault-workshop and challenge}.
\bibitem[{Lee et~al.(2013)}]{lee2013pseudo}
\bibinfo{author}{Lee, D.H.}, et~al., \bibinfo{year}{2013}.
\newblock \bibinfo{title}{Pseudo-label: The simple and efficient semi-supervised learning method for deep neural networks}, in: \bibinfo{booktitle}{Workshop on Challenges in Representation Learning, ICML}, p. \bibinfo{pages}{896}.
\bibitem[{Li et~al.(2018)Li, Chen, Qi, Dou, Fu and Heng}]{li2018h}
\bibinfo{author}{Li, X.}, \bibinfo{author}{Chen, H.}, \bibinfo{author}{Qi, X.}, \bibinfo{author}{Dou, Q.}, \bibinfo{author}{Fu, C.W.}, \bibinfo{author}{Heng, P.A.}, \bibinfo{year}{2018}.
\newblock \bibinfo{title}{H-denseunet: hybrid densely connected unet for liver and tumor segmentation from ct volumes}.
\newblock \bibinfo{journal}{IEEE Transactions on Medical Imaging} \bibinfo{volume}{37}, \bibinfo{pages}{2663--2674}.
\bibitem[{Liu et~al.(2023)Liu, Zhang, Chen, Xiao, Lu, Landman, Yuan, Yuille, Tang and Zhou}]{liu2023clip}
\bibinfo{author}{Liu, J.}, \bibinfo{author}{Zhang, Y.}, \bibinfo{author}{Chen, J.N.}, \bibinfo{author}{Xiao, J.}, \bibinfo{author}{Lu, Y.}, \bibinfo{author}{Landman, B.A.}, \bibinfo{author}{Yuan, Y.}, \bibinfo{author}{Yuille, A.}, \bibinfo{author}{Tang, Y.}, \bibinfo{author}{Zhou, Z.}, \bibinfo{year}{2023}.
\newblock \bibinfo{title}{Clip-driven universal model for organ segmentation and tumor detection}.
\newblock \bibinfo{journal}{arXiv preprint arXiv:2301.00785} .
\bibitem[{Liu and Zheng(2022)}]{liu2022context}
\bibinfo{author}{Liu, P.}, \bibinfo{author}{Zheng, G.}, \bibinfo{year}{2022}.
\newblock \bibinfo{title}{Context-aware voxel-wise contrastive learning for label efficient multi-organ segmentation}, in: \bibinfo{booktitle}{International Conference on Medical Image Computing and Computer-Assisted Intervention}, pp. \bibinfo{pages}{653--662}.
\bibitem[{Luo et~al.(2020)Luo, Yu, Chen, Liu, Wang, Xu and Heng}]{luo2020deep}
\bibinfo{author}{Luo, L.}, \bibinfo{author}{Yu, L.}, \bibinfo{author}{Chen, H.}, \bibinfo{author}{Liu, Q.}, \bibinfo{author}{Wang, X.}, \bibinfo{author}{Xu, J.}, \bibinfo{author}{Heng, P.A.}, \bibinfo{year}{2020}.
\newblock \bibinfo{title}{Deep mining external imperfect data for chest x-ray disease screening}.
\newblock \bibinfo{journal}{IEEE Transactions on Medical Imaging} \bibinfo{volume}{39}, \bibinfo{pages}{3583--3594}.
\bibitem[{Luo et~al.(2022)Luo, Hu, Liao, Zhai, Song, Wang and Zhang}]{luo2022scribble}
\bibinfo{author}{Luo, X.}, \bibinfo{author}{Hu, M.}, \bibinfo{author}{Liao, W.}, \bibinfo{author}{Zhai, S.}, \bibinfo{author}{Song, T.}, \bibinfo{author}{Wang, G.}, \bibinfo{author}{Zhang, S.}, \bibinfo{year}{2022}.
\newblock \bibinfo{title}{Scribble-supervised medical image segmentation via dual-branch network and dynamically mixed pseudo labels supervision}, in: \bibinfo{booktitle}{International Conference on Medical Image Computing and Computer-Assisted Intervention}, pp. \bibinfo{pages}{528--538}.
\bibitem[{Ma et~al.(2021)Ma, Zhang, Gu, Zhu, Ge, Zhang, An, Wang, Wang, Liu et~al.}]{ma2021abdomenct}
\bibinfo{author}{Ma, J.}, \bibinfo{author}{Zhang, Y.}, \bibinfo{author}{Gu, S.}, \bibinfo{author}{Zhu, C.}, \bibinfo{author}{Ge, C.}, \bibinfo{author}{Zhang, Y.}, \bibinfo{author}{An, X.}, \bibinfo{author}{Wang, C.}, \bibinfo{author}{Wang, Q.}, \bibinfo{author}{Liu, X.}, et~al., \bibinfo{year}{2021}.
\newblock \bibinfo{title}{Abdomenct-1k: Is abdominal organ segmentation a solved problem}.
\newblock \bibinfo{journal}{IEEE Transactions on Pattern Analysis and Machine Intelligence} .
\bibitem[{Van~der Maaten and Hinton(2008)}]{van2008visualizing}
\bibinfo{author}{Van~der Maaten, L.}, \bibinfo{author}{Hinton, G.}, \bibinfo{year}{2008}.
\newblock \bibinfo{title}{Visualizing data using t-sne.}
\newblock \bibinfo{journal}{Journal of Machine Learning Research} \bibinfo{volume}{9}.
\bibitem[{Nair et~al.(2020)Nair, Precup, Arnold and Arbel}]{nair2020exploring}
\bibinfo{author}{Nair, T.}, \bibinfo{author}{Precup, D.}, \bibinfo{author}{Arnold, D.L.}, \bibinfo{author}{Arbel, T.}, \bibinfo{year}{2020}.
\newblock \bibinfo{title}{Exploring uncertainty measures in deep networks for multiple sclerosis lesion detection and segmentation}.
\newblock \bibinfo{journal}{Medical image analysis} \bibinfo{volume}{59}, \bibinfo{pages}{101557}.
\bibitem[{Oh et~al.(2022)Oh, Kim and Kweon}]{oh2022daso}
\bibinfo{author}{Oh, Y.}, \bibinfo{author}{Kim, D.J.}, \bibinfo{author}{Kweon, I.S.}, \bibinfo{year}{2022}.
\newblock \bibinfo{title}{Daso: Distribution-aware semantics-oriented pseudo-label for imbalanced semi-supervised learning}, in: \bibinfo{booktitle}{Proceedings of the IEEE/CVF Conference on Computer Vision and Pattern Recognition}, pp. \bibinfo{pages}{9786--9796}.
\bibitem[{Oktay et~al.(2018)Oktay, Schlemper, Folgoc, Lee, Heinrich, Misawa, Mori, McDonagh, Hammerla, Kainz et~al.}]{oktay2018attention}
\bibinfo{author}{Oktay, O.}, \bibinfo{author}{Schlemper, J.}, \bibinfo{author}{Folgoc, L.L.}, \bibinfo{author}{Lee, M.}, \bibinfo{author}{Heinrich, M.}, \bibinfo{author}{Misawa, K.}, \bibinfo{author}{Mori, K.}, \bibinfo{author}{McDonagh, S.}, \bibinfo{author}{Hammerla, N.Y.}, \bibinfo{author}{Kainz, B.}, et~al., \bibinfo{year}{2018}.
\newblock \bibinfo{title}{Attention u-net: Learning where to look for the pancreas}.
\newblock \bibinfo{journal}{arXiv preprint arXiv:1804.03999} .
\bibitem[{Ouali et~al.(2020)Ouali, Hudelot and Tami}]{ouali2020semi}
\bibinfo{author}{Ouali, Y.}, \bibinfo{author}{Hudelot, C.}, \bibinfo{author}{Tami, M.}, \bibinfo{year}{2020}.
\newblock \bibinfo{title}{Semi-supervised semantic segmentation with cross-consistency training}, in: \bibinfo{booktitle}{Proceedings of the IEEE/CVF conference on computer vision and pattern recognition}, pp. \bibinfo{pages}{12674--12684}.
\bibitem[{Ronneberger et~al.(2015)Ronneberger, Fischer and Brox}]{ronneberger2015u}
\bibinfo{author}{Ronneberger, O.}, \bibinfo{author}{Fischer, P.}, \bibinfo{author}{Brox, T.}, \bibinfo{year}{2015}.
\newblock \bibinfo{title}{U-net: Convolutional networks for biomedical image segmentation}, in: \bibinfo{booktitle}{International Conference on Medical Image Computing and Computer-Assisted Intervention}, pp. \bibinfo{pages}{234--241}.
\bibitem[{Roth et~al.(2015)Roth, Lu, Farag, Shin, Liu, Turkbey and Summers}]{roth2015deeporgan}
\bibinfo{author}{Roth, H.R.}, \bibinfo{author}{Lu, L.}, \bibinfo{author}{Farag, A.}, \bibinfo{author}{Shin, H.C.}, \bibinfo{author}{Liu, J.}, \bibinfo{author}{Turkbey, E.B.}, \bibinfo{author}{Summers, R.M.}, \bibinfo{year}{2015}.
\newblock \bibinfo{title}{Deeporgan: Multi-level deep convolutional networks for automated pancreas segmentation}, in: \bibinfo{booktitle}{International Conference on Medical Image Computing and Computer-Assisted Intervention}, pp. \bibinfo{pages}{556--564}.
\bibitem[{Seibold et~al.(2022)Seibold, Rei{\ss}, Kleesiek and Stiefelhagen}]{seibold2022reference}
\bibinfo{author}{Seibold, C.M.}, \bibinfo{author}{Rei{\ss}, S.}, \bibinfo{author}{Kleesiek, J.}, \bibinfo{author}{Stiefelhagen, R.}, \bibinfo{year}{2022}.
\newblock \bibinfo{title}{Reference-guided pseudo-label generation for medical semantic segmentation}, in: \bibinfo{booktitle}{Proceedings of the AAAI conference on artificial intelligence}, pp. \bibinfo{pages}{2171--2179}.
\bibitem[{Shi et~al.(2021)Shi, Xiao, Chen and Zhou}]{shi2021marginal}
\bibinfo{author}{Shi, G.}, \bibinfo{author}{Xiao, L.}, \bibinfo{author}{Chen, Y.}, \bibinfo{author}{Zhou, S.K.}, \bibinfo{year}{2021}.
\newblock \bibinfo{title}{Marginal loss and exclusion loss for partially supervised multi-organ segmentation}.
\newblock \bibinfo{journal}{Medical Image Analysis} \bibinfo{volume}{70}, \bibinfo{pages}{101979}.
\bibitem[{Simpson et~al.(2019)Simpson, Antonelli, Bakas, Bilello, Farahani, Van~Ginneken, Kopp-Schneider, Landman, Litjens, Menze et~al.}]{simpson2019large}
\bibinfo{author}{Simpson, A.L.}, \bibinfo{author}{Antonelli, M.}, \bibinfo{author}{Bakas, S.}, \bibinfo{author}{Bilello, M.}, \bibinfo{author}{Farahani, K.}, \bibinfo{author}{Van~Ginneken, B.}, \bibinfo{author}{Kopp-Schneider, A.}, \bibinfo{author}{Landman, B.A.}, \bibinfo{author}{Litjens, G.}, \bibinfo{author}{Menze, B.}, et~al., \bibinfo{year}{2019}.
\newblock \bibinfo{title}{A large annotated medical image dataset for the development and evaluation of segmentation algorithms}.
\newblock \bibinfo{journal}{arXiv preprint arXiv:1902.09063} .
\bibitem[{Sohn et~al.(2020)Sohn, Berthelot, Carlini, Zhang, Zhang, Raffel, Cubuk, Kurakin and Li}]{sohn2020fixmatch}
\bibinfo{author}{Sohn, K.}, \bibinfo{author}{Berthelot, D.}, \bibinfo{author}{Carlini, N.}, \bibinfo{author}{Zhang, Z.}, \bibinfo{author}{Zhang, H.}, \bibinfo{author}{Raffel, C.A.}, \bibinfo{author}{Cubuk, E.D.}, \bibinfo{author}{Kurakin, A.}, \bibinfo{author}{Li, C.L.}, \bibinfo{year}{2020}.
\newblock \bibinfo{title}{Fixmatch: Simplifying semi-supervised learning with consistency and confidence}.
\newblock \bibinfo{journal}{Advances in Neural Information Processing Systems} \bibinfo{volume}{33}, \bibinfo{pages}{596--608}.
\bibitem[{Sun et~al.(2010)Sun, Zhang and Zhou}]{sun2010multi}
\bibinfo{author}{Sun, Y.Y.}, \bibinfo{author}{Zhang, Y.}, \bibinfo{author}{Zhou, Z.H.}, \bibinfo{year}{2010}.
\newblock \bibinfo{title}{Multi-label learning with weak label}, in: \bibinfo{booktitle}{Proceedings of the AAAI Conference on Artificial Intelligence}, pp. \bibinfo{pages}{593--598}.
\bibitem[{Tranheden et~al.(2021)Tranheden, Olsson, Pinto and Svensson}]{tranheden2021dacs}
\bibinfo{author}{Tranheden, W.}, \bibinfo{author}{Olsson, V.}, \bibinfo{author}{Pinto, J.}, \bibinfo{author}{Svensson, L.}, \bibinfo{year}{2021}.
\newblock \bibinfo{title}{Dacs: Domain adaptation via cross-domain mixed sampling}, in: \bibinfo{booktitle}{Proceedings of the IEEE/CVF Winter Conference on Applications of Computer Vision}, pp. \bibinfo{pages}{1379--1389}.
\bibitem[{Verelst et~al.(2023)Verelst, Rubenstein, Eichner, Tuytelaars and Berman}]{verelst2023spatial}
\bibinfo{author}{Verelst, T.}, \bibinfo{author}{Rubenstein, P.K.}, \bibinfo{author}{Eichner, M.}, \bibinfo{author}{Tuytelaars, T.}, \bibinfo{author}{Berman, M.}, \bibinfo{year}{2023}.
\newblock \bibinfo{title}{Spatial consistency loss for training multi-label classifiers from single-label annotations}, in: \bibinfo{booktitle}{Proceedings of the IEEE/CVF Winter Conference on Applications of Computer Vision}, pp. \bibinfo{pages}{3879--3889}.
\bibitem[{Wang et~al.(2019)Wang, Liew, Zou, Zhou and Feng}]{wang2019panet}
\bibinfo{author}{Wang, K.}, \bibinfo{author}{Liew, J.H.}, \bibinfo{author}{Zou, Y.}, \bibinfo{author}{Zhou, D.}, \bibinfo{author}{Feng, J.}, \bibinfo{year}{2019}.
\newblock \bibinfo{title}{Panet: Few-shot image semantic segmentation with prototype alignment}, in: \bibinfo{booktitle}{proceedings of the IEEE/CVF international conference on computer vision}, pp. \bibinfo{pages}{9197--9206}.
\bibitem[{Wang et~al.(2021)Wang, Peng and Zhang}]{wang2021uncertainty}
\bibinfo{author}{Wang, Y.}, \bibinfo{author}{Peng, J.}, \bibinfo{author}{Zhang, Z.}, \bibinfo{year}{2021}.
\newblock \bibinfo{title}{Uncertainty-aware pseudo label refinery for domain adaptive semantic segmentation}, in: \bibinfo{booktitle}{Proceedings of the IEEE/CVF international conference on computer vision}, pp. \bibinfo{pages}{9092--9101}.
\bibitem[{Wei et~al.(2023)Wei, Budd, Garcia-Peraza-Herrera, Dorent, Shi and Vercauteren}]{wei2023segmatch}
\bibinfo{author}{Wei, M.}, \bibinfo{author}{Budd, C.}, \bibinfo{author}{Garcia-Peraza-Herrera, L.C.}, \bibinfo{author}{Dorent, R.}, \bibinfo{author}{Shi, M.}, \bibinfo{author}{Vercauteren, T.}, \bibinfo{year}{2023}.
\newblock \bibinfo{title}{Segmatch: A semi-supervised learning method for surgical instrument segmentation}.
\newblock \bibinfo{journal}{arXiv preprint arXiv:2308.05232} .
\bibitem[{Wei et~al.(2016)Wei, Liang, Chen, Shen, Cheng, Feng, Zhao and Yan}]{wei2016stc}
\bibinfo{author}{Wei, Y.}, \bibinfo{author}{Liang, X.}, \bibinfo{author}{Chen, Y.}, \bibinfo{author}{Shen, X.}, \bibinfo{author}{Cheng, M.M.}, \bibinfo{author}{Feng, J.}, \bibinfo{author}{Zhao, Y.}, \bibinfo{author}{Yan, S.}, \bibinfo{year}{2016}.
\newblock \bibinfo{title}{Stc: A simple to complex framework for weakly-supervised semantic segmentation}.
\newblock \bibinfo{journal}{IEEE Transactions on Pattern Analysis and Machine Intelligence} \bibinfo{volume}{39}, \bibinfo{pages}{2314--2320}.
\bibitem[{Wu et~al.(2023)Wu, Li, Lin and Cheng}]{wu2023compete}
\bibinfo{author}{Wu, H.}, \bibinfo{author}{Li, X.}, \bibinfo{author}{Lin, Y.}, \bibinfo{author}{Cheng, K.T.}, \bibinfo{year}{2023}.
\newblock \bibinfo{title}{Compete to win: Enhancing pseudo labels for barely-supervised medical image segmentation}.
\newblock \bibinfo{journal}{IEEE Transactions on Medical Imaging} .
\bibitem[{Wu et~al.(2022)Wu, Wu, Wu, Ge and Cai}]{wu2022exploring}
\bibinfo{author}{Wu, Y.}, \bibinfo{author}{Wu, Z.}, \bibinfo{author}{Wu, Q.}, \bibinfo{author}{Ge, Z.}, \bibinfo{author}{Cai, J.}, \bibinfo{year}{2022}.
\newblock \bibinfo{title}{Exploring smoothness and class-separation for semi-supervised medical image segmentation}, in: \bibinfo{booktitle}{International conference on medical image computing and computer-assisted intervention}, \bibinfo{organization}{Springer}. pp. \bibinfo{pages}{34--43}.
\bibitem[{Xu et~al.(2020)Xu, Zhang, Ni, Li, Wang, Tian and Zhang}]{xu2020adversarial}
\bibinfo{author}{Xu, M.}, \bibinfo{author}{Zhang, J.}, \bibinfo{author}{Ni, B.}, \bibinfo{author}{Li, T.}, \bibinfo{author}{Wang, C.}, \bibinfo{author}{Tian, Q.}, \bibinfo{author}{Zhang, W.}, \bibinfo{year}{2020}.
\newblock \bibinfo{title}{Adversarial domain adaptation with domain mixup}, in: \bibinfo{booktitle}{Proceedings of the AAAI Conference on Artificial Intelligence}, pp. \bibinfo{pages}{6502--6509}.
\bibitem[{Xu et~al.(2022)Xu, Wang, Lu, Yu, Yan, Luo, Ma, Zheng and Tong}]{xu2022all}
\bibinfo{author}{Xu, Z.}, \bibinfo{author}{Wang, Y.}, \bibinfo{author}{Lu, D.}, \bibinfo{author}{Yu, L.}, \bibinfo{author}{Yan, J.}, \bibinfo{author}{Luo, J.}, \bibinfo{author}{Ma, K.}, \bibinfo{author}{Zheng, Y.}, \bibinfo{author}{Tong, R.K.y.}, \bibinfo{year}{2022}.
\newblock \bibinfo{title}{All-around real label supervision: Cyclic prototype consistency learning for semi-supervised medical image segmentation}.
\newblock \bibinfo{journal}{IEEE Journal of Biomedical and Health Informatics} \bibinfo{volume}{26}, \bibinfo{pages}{3174--3184}.
\bibitem[{Yang et~al.(2022)Yang, Zhuo, Qi, Shi and Gao}]{yang2022st++}
\bibinfo{author}{Yang, L.}, \bibinfo{author}{Zhuo, W.}, \bibinfo{author}{Qi, L.}, \bibinfo{author}{Shi, Y.}, \bibinfo{author}{Gao, Y.}, \bibinfo{year}{2022}.
\newblock \bibinfo{title}{St++: Make self-training work better for semi-supervised semantic segmentation}, in: \bibinfo{booktitle}{Proceedings of the IEEE/CVF conference on computer vision and pattern recognition}, pp. \bibinfo{pages}{4268--4277}.
\bibitem[{Yao et~al.(2021)Yao, Shi, Cao, Lu, Lu, Song, Jin, Xiao, Hou and Zhang}]{yao2021deepprognosis}
\bibinfo{author}{Yao, J.}, \bibinfo{author}{Shi, Y.}, \bibinfo{author}{Cao, K.}, \bibinfo{author}{Lu, L.}, \bibinfo{author}{Lu, J.}, \bibinfo{author}{Song, Q.}, \bibinfo{author}{Jin, G.}, \bibinfo{author}{Xiao, J.}, \bibinfo{author}{Hou, Y.}, \bibinfo{author}{Zhang, L.}, \bibinfo{year}{2021}.
\newblock \bibinfo{title}{Deepprognosis: Preoperative prediction of pancreatic cancer survival and surgical margin via comprehensive understanding of dynamic contrast-enhanced ct imaging and tumor-vascular contact parsing}.
\newblock \bibinfo{journal}{Medical Image Analysis} \bibinfo{volume}{73}, \bibinfo{pages}{102150}.
\bibitem[{Ye et~al.(2023)Ye, Xie, Zhang, Chen and Xia}]{ye2023uniseg}
\bibinfo{author}{Ye, Y.}, \bibinfo{author}{Xie, Y.}, \bibinfo{author}{Zhang, J.}, \bibinfo{author}{Chen, Z.}, \bibinfo{author}{Xia, Y.}, \bibinfo{year}{2023}.
\newblock \bibinfo{title}{Uniseg: A prompt-driven universal segmentation model as well as a strong representation learner}.
\newblock \bibinfo{journal}{arXiv preprint arXiv:2304.03493} .
\bibitem[{Yun et~al.(2019)Yun, Han, Oh, Chun, Choe and Yoo}]{yun2019cutmix}
\bibinfo{author}{Yun, S.}, \bibinfo{author}{Han, D.}, \bibinfo{author}{Oh, S.J.}, \bibinfo{author}{Chun, S.}, \bibinfo{author}{Choe, J.}, \bibinfo{author}{Yoo, Y.}, \bibinfo{year}{2019}.
\newblock \bibinfo{title}{Cutmix: Regularization strategy to train strong classifiers with localizable features}, in: \bibinfo{booktitle}{Proceedings of the IEEE/CVF International Conference on Computer Vision}, pp. \bibinfo{pages}{6023--6032}.
\bibitem[{Zhang et~al.(2024)Zhang, Lin, Tang and Cheng}]{zhang2024cae}
\bibinfo{author}{Zhang, D.}, \bibinfo{author}{Lin, Y.}, \bibinfo{author}{Tang, J.}, \bibinfo{author}{Cheng, K.T.}, \bibinfo{year}{2024}.
\newblock \bibinfo{title}{Cae-great: Convolutional-auxiliary efficient graph reasoning transformer for dense image predictions}.
\newblock \bibinfo{journal}{International Journal of Computer Vision} \bibinfo{volume}{132}, \bibinfo{pages}{1502--1520}.
\bibitem[{Zhang et~al.(2020a)Zhang, Zhang, Tang, Hua and Sun}]{zhang2020causal}
\bibinfo{author}{Zhang, D.}, \bibinfo{author}{Zhang, H.}, \bibinfo{author}{Tang, J.}, \bibinfo{author}{Hua, X.S.}, \bibinfo{author}{Sun, Q.}, \bibinfo{year}{2020}a.
\newblock \bibinfo{title}{Causal intervention for weakly-supervised semantic segmentation}.
\newblock \bibinfo{journal}{Advances in Neural Information Processing Systems} \bibinfo{volume}{33}, \bibinfo{pages}{655--666}.
\bibitem[{Zhang et~al.(2020b)Zhang, Zhang, Tang, Wang, Hua and Sun}]{zhang2020feature}
\bibinfo{author}{Zhang, D.}, \bibinfo{author}{Zhang, H.}, \bibinfo{author}{Tang, J.}, \bibinfo{author}{Wang, M.}, \bibinfo{author}{Hua, X.}, \bibinfo{author}{Sun, Q.}, \bibinfo{year}{2020}b.
\newblock \bibinfo{title}{Feature pyramid transformer}, in: \bibinfo{booktitle}{Computer Vision--ECCV 2020: 16th European Conference, Glasgow, UK, August 23--28, 2020, Proceedings, Part XXVIII 16}, \bibinfo{organization}{Springer}. pp. \bibinfo{pages}{323--339}.
\bibitem[{Zhang et~al.(2021a)Zhang, Xie, Xia and Shen}]{zhang2021dodnet}
\bibinfo{author}{Zhang, J.}, \bibinfo{author}{Xie, Y.}, \bibinfo{author}{Xia, Y.}, \bibinfo{author}{Shen, C.}, \bibinfo{year}{2021}a.
\newblock \bibinfo{title}{Dodnet: Learning to segment multi-organ and tumors from multiple partially labeled datasets}, in: \bibinfo{booktitle}{Proceedings of the IEEE/CVF Conference on Computer Vision and Pattern Recognition}, pp. \bibinfo{pages}{1195--1204}.
\bibitem[{Zhang and Zhuang(2022)}]{zhang2022deep}
\bibinfo{author}{Zhang, K.}, \bibinfo{author}{Zhuang, X.}, \bibinfo{year}{2022}.
\newblock \bibinfo{title}{Deep compatible learning for partially-supervised medical image segmentation}.
\newblock \bibinfo{journal}{arXiv preprint arXiv:2206.09148} .
\bibitem[{Zhang et~al.(2022)Zhang, Abdelfattah, Song and Wang}]{zhang2022effective}
\bibinfo{author}{Zhang, X.}, \bibinfo{author}{Abdelfattah, R.}, \bibinfo{author}{Song, Y.}, \bibinfo{author}{Wang, X.}, \bibinfo{year}{2022}.
\newblock \bibinfo{title}{An effective approach for multi-label classification with missing labels}.
\newblock \bibinfo{journal}{arXiv preprint arXiv:2210.13651} .
\bibitem[{Zhang et~al.(2021b)Zhang, Liu, Ou, Zeng, Xiong, Yu, Liu and Ye}]{zhang2021carvemix}
\bibinfo{author}{Zhang, X.}, \bibinfo{author}{Liu, C.}, \bibinfo{author}{Ou, N.}, \bibinfo{author}{Zeng, X.}, \bibinfo{author}{Xiong, X.}, \bibinfo{author}{Yu, Y.}, \bibinfo{author}{Liu, Z.}, \bibinfo{author}{Ye, C.}, \bibinfo{year}{2021}b.
\newblock \bibinfo{title}{Carvemix: A simple data augmentation method for brain lesion segmentation}, in: \bibinfo{booktitle}{International Conference on Medical Image Computing and Computer-Assisted Intervention}, pp. \bibinfo{pages}{196--205}.
\bibitem[{Zhou et~al.(2021)Zhou, Ye and Zhan}]{zhou2021learning}
\bibinfo{author}{Zhou, D.W.}, \bibinfo{author}{Ye, H.J.}, \bibinfo{author}{Zhan, D.C.}, \bibinfo{year}{2021}.
\newblock \bibinfo{title}{Learning placeholders for open-set recognition}, in: \bibinfo{booktitle}{Proceedings of the IEEE/CVF Conference on Computer Vision and Pattern Recognition}, pp. \bibinfo{pages}{4401--4410}.
\bibitem[{Zhou et~al.(2022)Zhou, Wang, Konukoglu and Van~Gool}]{zhou2022rethinking}
\bibinfo{author}{Zhou, T.}, \bibinfo{author}{Wang, W.}, \bibinfo{author}{Konukoglu, E.}, \bibinfo{author}{Van~Gool, L.}, \bibinfo{year}{2022}.
\newblock \bibinfo{title}{Rethinking semantic segmentation: A prototype view}, in: \bibinfo{booktitle}{Proceedings of the IEEE/CVF Conference on Computer Vision and Pattern Recognition}, pp. \bibinfo{pages}{2582--2593}.
\bibitem[{Zhou et~al.(2019)Zhou, Li, Bai, Wang, Chen, Han, Fishman and Yuille}]{zhou2019prior}
\bibinfo{author}{Zhou, Y.}, \bibinfo{author}{Li, Z.}, \bibinfo{author}{Bai, S.}, \bibinfo{author}{Wang, C.}, \bibinfo{author}{Chen, X.}, \bibinfo{author}{Han, M.}, \bibinfo{author}{Fishman, E.}, \bibinfo{author}{Yuille, A.L.}, \bibinfo{year}{2019}.
\newblock \bibinfo{title}{Prior-aware neural network for partially-supervised multi-organ segmentation}, in: \bibinfo{booktitle}{Proceedings of the IEEE/CVF International Conference on Computer Vision}, pp. \bibinfo{pages}{10672--10681}.
\bibitem[{Zhu et~al.(2023)Zhu, Zhu, Yu and Li}]{zhu2023progressive}
\bibinfo{author}{Zhu, R.}, \bibinfo{author}{Zhu, R.}, \bibinfo{author}{Yu, X.}, \bibinfo{author}{Li, S.}, \bibinfo{year}{2023}.
\newblock \bibinfo{title}{Progressive mix-up for few-shot supervised multi-source domain transfer}.
\newblock \bibinfo{journal}{The Eleventh International Conference on Learning Representations} .
\bibitem[{Zou et~al.(2020)Zou, Zhang, Zhang, Li, Bian, Huang and Pfister}]{zou2020pseudoseg}
\bibinfo{author}{Zou, Y.}, \bibinfo{author}{Zhang, Z.}, \bibinfo{author}{Zhang, H.}, \bibinfo{author}{Li, C.L.}, \bibinfo{author}{Bian, X.}, \bibinfo{author}{Huang, J.B.}, \bibinfo{author}{Pfister, T.}, \bibinfo{year}{2020}.
\newblock \bibinfo{title}{Pseudoseg: Designing pseudo labels for semantic segmentation}.
\newblock \bibinfo{journal}{arXiv preprint arXiv:2010.09713} .

\end{thebibliography}
\end{document}